\pdfoutput=1
\documentclass[11pt]{article}

\usepackage{acl}

\usepackage{times}
\usepackage{latexsym}

\usepackage[T1]{fontenc}
\usepackage[utf8]{inputenc}

\usepackage{microtype}

\usepackage{inconsolata}

\usepackage{graphicx}

\usepackage{amsmath}
\usepackage{xspace}
\newcommand{\lib}{\textsc{MultiPun}\xspace}

\usepackage[textwidth=3.5cm]{todonotes}
\usepackage{bm}

\usepackage{amssymb}
\usepackage{algorithm}
\usepackage{algorithmic}

\usepackage{fontawesome}
\usepackage{multirow} 
\usepackage{colortbl} 
\usepackage{booktabs}
\definecolor{darkgreen}{RGB}{0, 100, 0}
\usepackage{siunitx}
\usepackage{makecell}
\usepackage{tcolorbox}
\tcbuselibrary{listings,breakable}
\usepackage{amsthm}
\definecolor{mypink}{RGB}{255,231,226}



\usepackage{soul}
\usepackage{xcolor} % For defining colors
\usepackage{colortbl}
\usepackage{tcolorbox}
\usepackage{pgf}

\newtcolorbox{mybox3}[1]{colbacktitle=white,coltitle=black,colback=white,colframe=black,fonttitle=\bfseries,fontupper=\small,title=#1,leftupper=0.5em,rightupper=0.5em,boxrule=1.0pt}

\newtcolorbox{PromptBox}[1]{
    colback=gray!10,
    colframe=black!60,
    title=#1,
    fonttitle=\bfseries,
    breakable,
    boxrule=0.5pt,
    arc=2mm,
    fontupper=\footnotesize, % 设置内容字体为 \footnotesize
}

\usepackage{enumitem}

\tcbset{
    fontupper=\ttfamily\small, %\normalsize, % \scriptsize,    % Monospace font for upper part
    fontlower=\ttfamily\small, % \normalsize, % \scriptsize,    % Monospace font for lower part
    halign=left,    % This sets left alignment for the box content
    boxsep=1mm, left=1mm, right=1mm, top=1mm, bottom=1mm
}

\title{``I See What You Did There'': Can Large Vision-Language Models Understand Multimodal Puns?}

\author{
   ~~Naen Xu$^{1}$,
   ~~Jiayi Sheng$^{2}$,
   ~~Changjiang Li$^{3}$,
   ~~Chunyi Zhou$^{1}$, 
   ~~Yuyuan Li$^{4}$, \\
   ~~\textbf{Tianyu Du}$^{1,5}$\thanks{$\quad$ Corresponding Author.}\textbf{,}
   ~~\textbf{Jun Wang}$^{6}$\footnotemark[1]\textbf{,}
   ~~\textbf{Zhihui Fu}$^{6}$\textbf{,}
   ~~\textbf{Jinbao Li}$^{7}$\textbf{,}
   ~~\textbf{Shouling Ji}$^{1}$\\ 
   $^{1}$Zhejiang University, $^{2}$Beihang University, \\
   $^{3}$Palo Alto Networks, $^{4}$Hangzhou Dianzi University,  \\
   $^{5}$Ningbo Global Innovation Center, Zhejiang University,\\
   $^{6}$OPPO Research Institute, $^{7}$Qilu University of Technology \\
   \texttt{\{xunaen, zjradty\}@zju.edu.cn, junwang.lu@gmail.com} % \\
}

\begin{document}
\maketitle

% \renewcommand{\thefootnote}{\fnsymbol{footnote}}
% \footnotetext[1]{Corresponding author.}
% \renewcommand{\thefootnote}{\arabic{footnote}}

\begin{abstract}
Puns are a common form of rhetorical wordplay that exploits polysemy and phonetic similarity to create humor. 
In multimodal puns, visual and textual elements synergize to ground the literal sense and evoke the figurative meaning simultaneously.
Although Vision-Language Models (VLMs) are widely used in multimodal understanding and generation, their ability to understand puns has not been systematically studied due to a scarcity of rigorous benchmarks. 
To address this, we first propose a multimodal pun generation pipeline. We then introduce \textsc{MultiPun}, a dataset comprising diverse types of puns alongside adversarial non-pun distractors. 
Our evaluation reveals that most models struggle to distinguish genuine puns from these distractors. 
Moreover, we propose both prompt-level and model-level strategies to enhance pun comprehension, with an average improvement of 16.5\% in F1 scores. 
Our findings provide valuable insights for developing future VLMs that master the subtleties of human-like humor via cross-modal reasoning.
\end{abstract}

% \vspace{-3px}
\section{Introduction}
% \vspace{-1px}
\label{introduction}

Puns, also known as paronomasia in linguistics, are a form of wordplay that exploits multiple meanings of a term or similar-sounding words to create humor~\cite{miller2015automatic,kao2016computational}. Interpreting multimodal puns requires resolving a complex visual synthesis beyond simple image captioning: the image fuses a literal object with a metaphorical context, while the text forces a dual interpretation by unifying the object's visual identity with its behavioral state. Compared to other forms of humor like jokes~\cite{dynel2009beyond} or comedies~\cite{stott2014comedy}, puns are structurally simpler and possess more precise linguistic definitions~\cite{hempelmann2008computational,attardo2018universals}. These qualities make them an ideal testbed for evaluating multimodal reasoning in Vision-Language Models (VLMs)~\cite{team2023gemini}.

\begin{figure}[t]
\begin{center}
\centerline{\includegraphics[width=\linewidth]{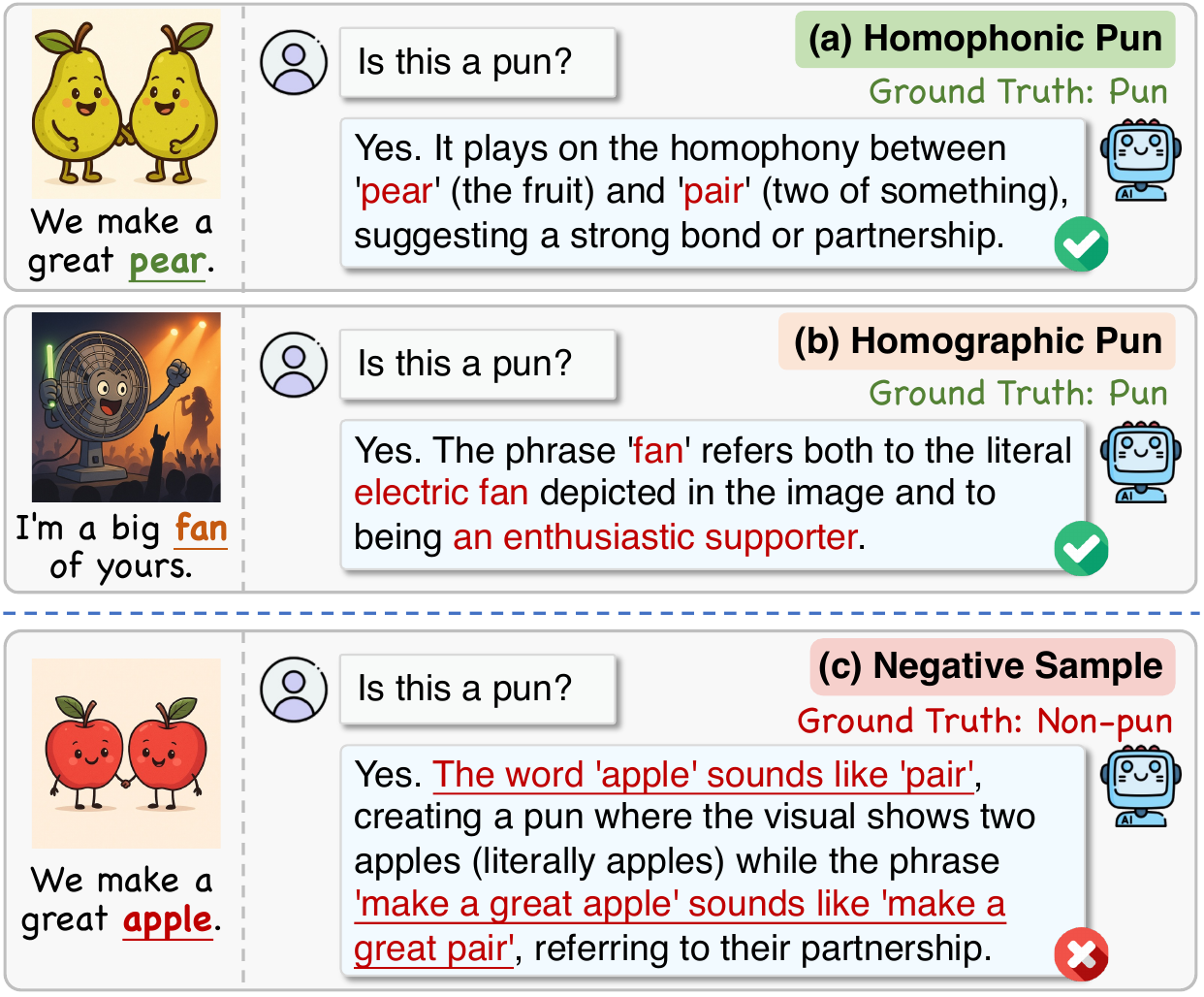}}
\vspace{-5px}
\caption{
The recognition of multimodal pun examples from \lib. (a) A pun relying on phonetic similarity (``pear'' and ``pair''). (b) A pun based on word polysemy (double meaning of ``fan''). (c) A negative sample illustrating a false positive case, where the model mistakenly interprets a non-pun as pun.}

\label{fig:part}
\end{center}
\vspace{-25px}
\end{figure}

Consider the examples in Figure~\ref{fig:part}. 
Figure~\ref{fig:part}(a) depicts two pears (literal objects) holding hands like a romantic couple (figurative behavior). 
The caption ``We make a great \textit{pear}'' exploits the sound similarity of ``pear'' to ``pair''. 
The humor emerges by connecting the visual intimacy (holding hands) with the auditory implication of being a romantic ``pair''. 
Similarly, Figure~\ref{fig:part}(b) also relies on the double meanings of the same word. The caption ``I'm a big \textit{fan} of yours'' uses the polysemy of ``fan'' (cooling device vs. enthusiast). Notably, the image depicts an industrial fan (literal object), cheering with a glow stick (figurative behavior). 
Crucially, Figure~\ref{fig:part}(c) presents a negative example. The image still depicts intimate fruits (apples) and the sentence structure remains identical, but the phonetic connection to ``pair'' is broken. A robust model should recognize it as non-pun, whereas existing models might mistakenly interpret it as a pun.

Recent studies on pun detection~\cite{zhou-etal-2020-boating}, explanation~\cite{zangari-etal-2025-pun}, and generation~\cite{xu-etal-2024-good} face three critical limitations.
(\textit{i}) \textbf{Unimodal confinement.}
Prior research predominantly targets textual puns~\cite{miller-etal-2017-semeval}, overlooking the complex cross-modal interplay where visual modality can also cause ambiguity.
(\textit{ii}) \textbf{Deficiencies in multimodal benchmarks.} 
Existing multimodal efforts~\cite{xu-etal-2025-punmemecn} lack detailed pun categorization and non-puns as negative samples. This positive-only approach prevents us from knowing whether models truly understand the pun or just superficially link playful visual scenes with humor.
(\textit{iii}) \textbf{Conflation of preference and comprehension.} 
Previous evaluations~\cite{xu-etal-2025-punmemecn,zangari-etal-2025-pun} rely on single-sided querying (e.g., ``Is this a pun?''), failing to separate true reasoning from the model's affirmative language bias~\cite{zhuang2024beyond}. 
To address these gaps, we summarize three research questions (RQs):
\begin{itemize}[nosep,leftmargin=11pt]
    \item \textbf{RQ\textsubscript{1}} -- How effectively can VLMs recognize multimodal puns against non-puns?
    \item \textbf{RQ\textsubscript{2}} -- To what extent can VLMs explain puns?
    \item \textbf{RQ\textsubscript{3}} -- How can we enhance VLMs' understanding of puns?
\end{itemize}

To assess the abilities of VLMs in multimodal pun understanding, we propose \textsc{MultiPun}, a linguistically grounded multimodal benchmark with both pun and non-pun samples.
To address \textbf{RQ\textsubscript{1}}, we assess models' performance in pun detection, localization, and explanation tasks. % We ask the same question in both direct and reverse forms, analyzing responses together to disentangle model preference from true pun understanding.
For \textbf{RQ\textsubscript{2}}, we employ both a fine-grained pun component verification and a coarse-grained explanation pairwise comparison to assess VLMs’ comprehension of puns.
Finally, to answer \textbf{RQ\textsubscript{3}}, we propose prompt-level and model-level strategies to enhance VLMs' understanding of puns. 
In summary, our contributions are as follows:
\begin{itemize}[nosep,leftmargin=11pt]
    \item We introduce the multimodal pun generation pipeline and propose \textsc{MultiPun}, a benchmark containing 445 puns and 890 non-puns to evaluate VLMs' understanding of puns.
    \item  We design three pun detection, localization, and explanation tasks, and find that most VLMs superficially connect puns to common language patterns rather than truly understand them.
    \item We provide prompt-level method Pun-CoT and model-level method Pun-Tuning to enhance VLMs' understanding of puns, resulting in an average increase of 16.5\% in F1 scores.
\end{itemize}

% \vspace{-3px}
\section{Related Work}
\vspace{-2px}
\label{related_work}

\paragraph{Textual pun understanding.}
Puns are a linguistic art form that relies on phonological or semantic ambiguity. 
Early research primarily focuses on curating textual pun collections from web sources~\cite{miller-etal-2017-semeval}.
The field gained momentum with SemEval-2017 Task 7~\cite{miller-etal-2017-semeval}, which established benchmarks for pun detection and location. 
Recently, researchers have used Large Language Models (LLMs) to advance the detection~\cite{zou-lu-2019-joint,zhou-etal-2020-boating}, explanation~\cite{sun-etal-2022-expunations}, and generation~\cite{yu-etal-2020-homophonic} of puns.
However, these studies are confined to the textual modality, ignoring the cognitive complexity of multimodal ambiguity. Our work extends this by integrating the visual modality as an essential component for resolving ambiguity.

\paragraph{Multimodal humor and pun understanding.}
Understanding visual humor is crucial for assessing multimodal reasoning in VLMs. While there is growing interest in memes~\cite{liu2024ii,xu-etal-2025-punmemecn}, sarcasm~\cite{wang2025can}, comics~\cite{hu2024cracking} and Chinese pun rebus~\cite{zhang2025creating}, research on multimodal puns is limited. Existing datasets lack fine-grained linguistic categorization, failing to distinguish between phonological and semantic strategies. More critically, most benchmarks evaluate models solely on puns without rigorous negative samples~\cite{xu-etal-2025-punmemecn,chung-etal-2024-visual}. This makes it hard to determine whether models truly understand cross-modal alignment or merely generate hallucinatory humor. Our work bridges this gap with a benchmark including adversarial negatives.

\begin{figure*}[t]
\begin{center}
\centerline{\includegraphics[width=\linewidth]{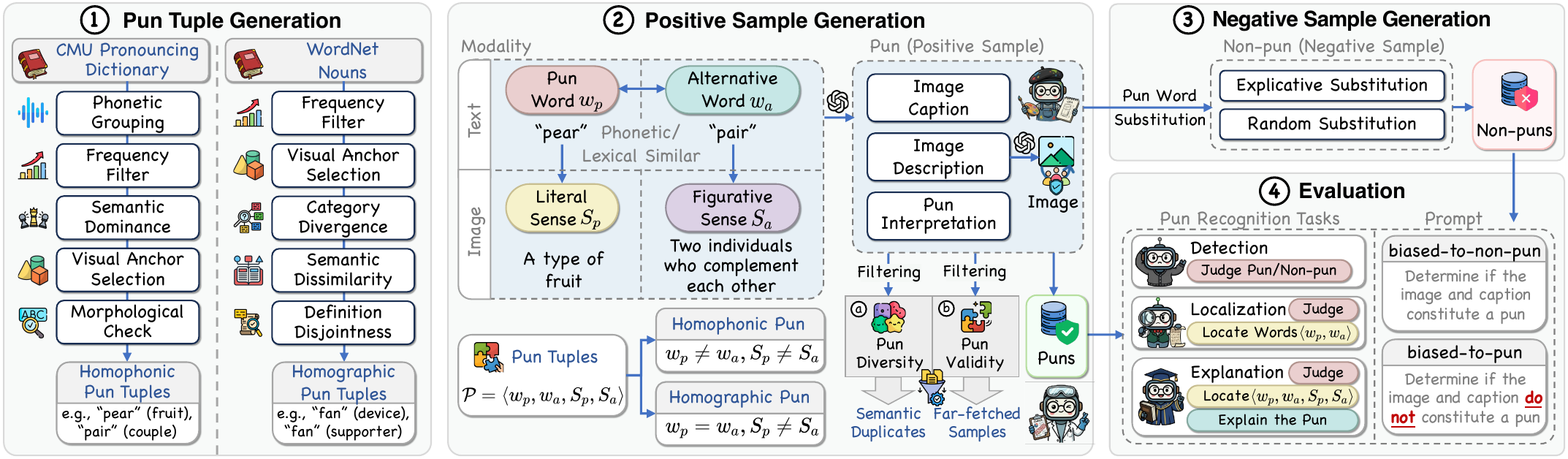}}
\vspace{-5px}
\caption{Overview of the \lib construction pipeline. Our pipeline generates both pun and non-pun samples.}
\label{fig:pipeline}
\end{center}
\vspace{-25px}
\end{figure*}

\vspace{-2px}
\section{\lib}
\vspace{-2px}
\label{methods}
\textsc{MultiPun} is a multimodal benchmark with 445 puns (homophonic and homographic, Section~\ref{sec:preliminary}) and 890 non-pun distractors from two substitution strategies. Figure~\ref{fig:pipeline} shows the construction pipeline (Section~\ref{sec:construction}). We introduce an evaluation suite for pun detection, localization, and explanation (Section~\ref{sec:task}) to assess VLM performance.

\vspace{-2px}
\subsection{Preliminary}
\vspace{-2px}
\label{sec:preliminary}
We focus on two main types of multimodal puns: \textit{homophonic puns} and \textit{homographic puns}~\cite{miller-etal-2017-semeval}.
We formalize a multimodal pun instance as a tuple $\mathcal{P} = \langle w_p, w_a, S_p, S_a \rangle$ following \citet{xu-etal-2024-good}. 
Here, $w_p$ denotes the \textit{pun word} in the image caption, and $w_a$ represents the \textit{alternative word}. Crucially, the image fuses two semantics: $S_p$ is the literal concrete object corresponding to the meaning of $w_p$, and $S_a$ is the figurative behavior or state associated with $w_a$.
\begin{itemize}[nosep,leftmargin=11pt]
    \item \textbf{Homophonic Pun:} This category uses the sound similarity between the $w_p$ in the caption and $w_a$, which differ in spelling and meaning~\cite{attardo2024linguistic}. % The image shows $w_p$ with meaning $S_p$, while depicting the behavior or state of $w_a$ with meaning $S_a$. 
    For instance, Figure~\ref{fig:part}(a) shows pears ($S_p$) holding hands like a couple ($S_a$), hinting at the ``pair'' ($w_a$), phonetically triggered by ``We make a great \textit{pear}'' ($w_p$).

    \item \textbf{Homographic Pun:} This category exploits the dual meaning of homographs~\cite{attardo2024linguistic}, where $w_p$ and $w_a$ are spelled the same but have different meanings. % The image depicts the subject that embodies the concrete sense ($S_p$) while simultaneously fulfilling the role of the abstract or figurative sense ($S_a$). 
    For example, in Figure~\ref{fig:part}(b), ``fan'' serves as both the cooling device ($w_p$) and the enthusiast ($w_a$). The visual subject physically embodies the device ($S_p$) while functionally enacting the cheering behavior ($S_a$).

\end{itemize}
% The main challenge for VLMs lies in cross-modal reasoning: recognizing that a single visual entity is a composite of two distinct semantic concepts ($S_p$ and $S_a$), linked by the linguistic trigger ($w_p$).

\vspace{-1px}
\subsection{Dataset Construction}
\label{sec:construction}
\vspace{-1px}

As shown in Figure~\ref{fig:pipeline}, we construct the \textsc{MultiPun} benchmark using the following pipeline: pun tuple generation, positive sample generation, negative sample generation, and evaluation.

\vspace{-1px}
\subsubsection{Pun Tuples Generation}

\paragraph{Homophonic Puns.} 
We retrieve word pairs $w_p$ and $w_a$ with identical pronunciation but distinct spellings with the following steps:
(\textit{i}) \textit{Phonetic Grouping:} Use the CMU Pronouncing Dictionary~\cite{cmu_dict} to find word pairs with identical pronunciation.
(\textit{ii}) \textit{Frequency Filter:} Apply a Zipf frequency threshold ($> 3.0$) to ensure words are commonly used.
(\textit{iii}) \textit{Semantic Dominance:} Select the top-3 most frequent synsets in WordNet~\cite{miller-1992-wordnet} to prioritize primary meanings.
(\textit{iv}) \textit{Visual Anchor Selection:} Keep concrete nouns in visually depictable categories (\textit{e.g.}, \textit{noun.animal}, \textit{noun.artifact}) so that $w_p$ can be clearly illustrated.
(\textit{v}) \textit{Morphological Check:} Use lemmatization checks to remove trivial variants, ensuring $w_p$ and $w_a$ are distinct lemmas.

\paragraph{Homographic Puns.} 
We retrieve word $w_p$ with two different meanings $S_p$ and $S_a$ with the following steps:
(\textit{i}) \textit{Frequency Filter:} Select nouns with a Zipf frequency over 3.8 and choose their top-3 WordNet~\cite{miller-1992-wordnet} synsets.
(\textit{ii}) \textit{Visual Anchor Selection:} Keep candidates with a concrete sense ($S_p$) in visually depictable noun categories (\textit{e.g.}, \textit{noun.animal}, \textit{noun.artifact}) so that $w_p$ can be clearly illustrated.
(\textit{iii}) \textit{Category Divergence:} Ensure $S_a$ is in a different lexical file from $S_p$.
(\textit{iv}) \textit{Semantic Dissimilarity:} Require low path similarity ($< 0.1$) and reject pairs where both in natural categories (\textit{e.g.}, \textit{noun.plant}, \textit{noun.animal}), avoiding part-whole metonymy (e.g., apple tree vs. apple fruit).
(\textit{v}) \textit{Definition Disjointness:} Remove synsets with definitions containing the target word, avoiding circular meanings (e.g., rejecting the ``ball game'' sense of \textit{baseball} includes ``ball'').

\noindent After filtering, we obtain a set of pun tuples $\mathcal{P} = \langle w_p, w_a, S_p, S_a \rangle$ as seeds for sample generation.

\begin{figure*}[t]
\begin{center}
\centerline{\includegraphics[width=\linewidth]{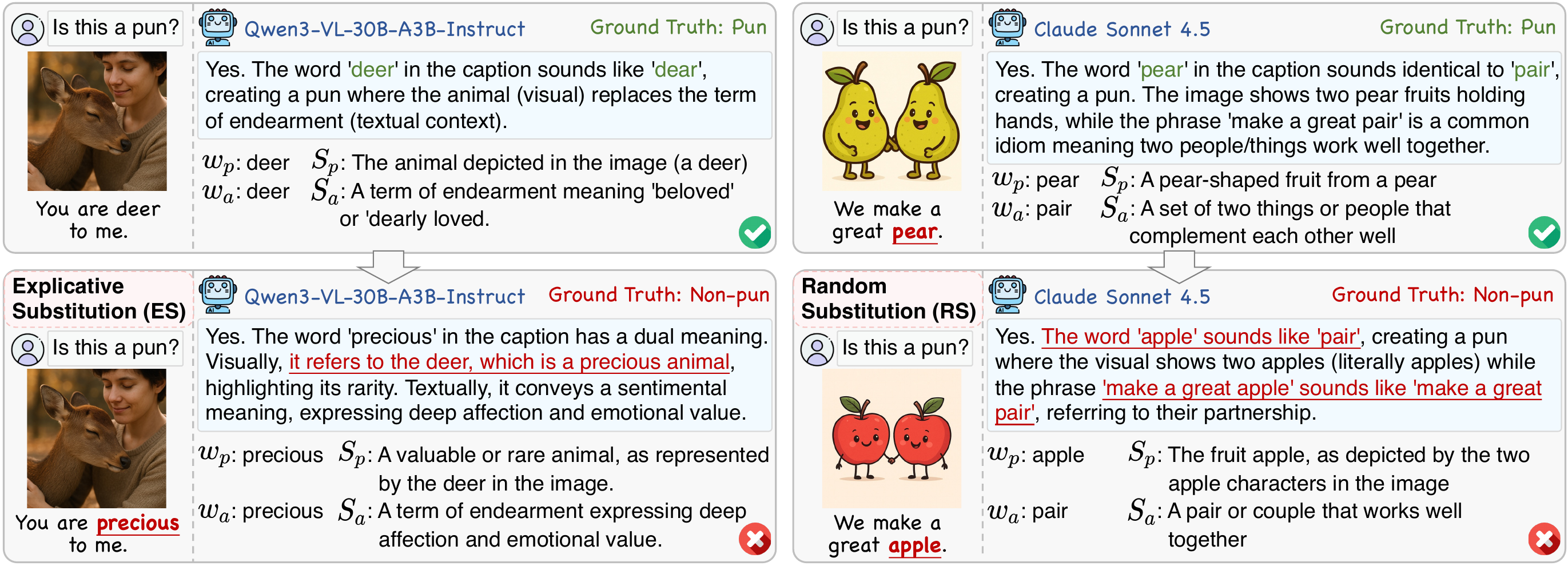}}
\vspace{-5px}
\caption{Examples of adversarial negative samples.} %  (a) Original Homophonic Pun with ``pear/pair'': the pun word unifies the fruit identity with romantic behavior. (b) Explicative Substitution: explicitly states ``romantic couple,'' eliminating linguistic ambiguity. (c) Random Substitution: replaces with ``banana,'' preserving anthropomorphization but removing the pun structure (``banana'' has no dual meaning related to romance).
\label{fig:negative}
\end{center}
\vspace{-25px}
\end{figure*}

\subsubsection{Positive Sample Generation}

\paragraph{Generation.}
Based on the pun tuples from the previous step, we employ \texttt{GPT-4o} to create multimodal samples. Specifically, the model is prompted to generate three distinct components for each tuple: (\textit{i}) an image caption containing the pun word $w_p$, (\textit{ii}) an image description detailed enough to guide the text-to-image generation, and (\textit{iii}) a pun interpretation explaining the ambiguity. The image description is subsequently fed into the image generator \texttt{GPT-image-1} to create the visual scene. We manually verify image-description alignment and refine prompts to regenerate images when mismatches occur. The visual scene grounds the object's identity in the literal sense ($S_p$) while enacting its behavior in the figurative sense ($S_a$).

\paragraph{Filtering.}
We use the following filtering steps:
(\textit{i}) \textbf{Diversity Filtering:} Embedding-based filtering model \texttt{text-embedding-3-large}~\cite{openai_embedding_v3} removes highly similar samples to eliminate redundancy (see Appendix~\ref{app:diversity_filtering} for the algorithm).
(\textit{ii}) \textbf{Validity Filtering:} We employ human-in-the-loop quality control to final verification (details in Appendix~\ref{appendix:human_verification}). We discard \textit{far-fetched samples} where the connection between the image and the caption is insufficient to form a recognizable pun.

\subsubsection{Negative Sample Generation}
\label{sec:negative_construction}
\vspace{-2px}

% Existing benchmarks often suffer from a positive-only bias, failing to detect whether VLMs recognize superficial cues or overfit to specific pun structures without fully understanding their meanings.
% % 你在干什么？？？？？？？？？？喂
% % shishuo都下线了 我给你笑死了!!他在看晚上吃啥 
% % 你是不是都偷偷润色过了
% % 我感觉没啥好改的
% % 没有啊啊啊bb
% To address this, we create two types of adversarial negative samples to disrupt puns while preserving surface coherence, and then verify their quality through human checks. 
To mitigate the positive-only bias in existing benchmarks and distinguish genuine comprehension from superficial overfitting, we construct adversarial negatives that disrupt the pun mechanism while maintaining surface coherence. We employ two primary disruption strategies:
\begin{itemize}[nosep,leftmargin=11pt]
    \item \textbf{Explicative Substitution (ES):} 
    This variant resolves the linguistic ambiguity by replacing the pun word $w_p$ with a direct description of the behavioral meaning $S_a$.
    \item \textbf{Random Substitution (RS):} 
    This variant replaces $w_p$ with a semantically unrelated entity (\textit{e.g.}, ``chair'', ``apple''), and creates a new image where a new entity performs the original action.
\end{itemize}
%  by substituting pun words with non-pun words using GPT-4o

\subsubsection{Evaluation Tasks}
\label{sec:task}
\vspace{-2px}

To systematically assess VLMs' capabilities in multimodal pun comprehension, we design a progressive evaluation suite comprising three tasks.
\begin{itemize}[nosep,leftmargin=11pt]
    \item \textbf{Detection} asks for binary judgment (pun or not) without definitions or guidance.
    \item \textbf{Localization} requires first judging and explicitly identifying words $w_p$ and $w_a$.
    \item \textbf{Explanation} requires judging, providing a rationale that explains why it's a pun, and extracting the full tuple $\langle w_p, w_a, S_p, S_a \rangle$.  
\end{itemize}
To separate true reasoning from the model’s affirmative language bias~\cite{zhuang2024beyond,xu-etal-2024-good}, we ask the same question twice in both direct and opposite form: (\textit{i}) a \textit{biased-to-pun} prompt that asks whether the given multimodal context is a pun, and (\textit{ii}) a \textit{biased-to-non-pun} prompt that asks whether the given multimodal context is not a pun.\footnote{All prompts are provided in Appendix~\ref{appendix:eval_prompts}.}

\subsection{Experimental Setup}
\vspace{-2px}
\paragraph{Models.}
We evaluate 11 representative VLMs on \lib across three tasks to evaluate their understanding of the puns, including GPT~\cite{openai2024gpt5}, Gemini~\cite{comanici2025gemini}, Claude~\cite{anthropic2024claudesonnet},
Qwen~\cite{Qwen3-VL}, LLaVA~\cite{liu2023visual} series.\footnote{Detailed settings of VLMs are given in Appendix~\ref{app:models}.}
% InternVL3.5~\cite{wang2025internvl3}, DeepSeek-VL2~\cite{wu2024deepseek}, 

\paragraph{Metrics.}
We use two categories of metrics to evaluate model performance. 
(\textit{i}) \textbf{Pun Recognition.} For all tasks (detection, localization, and explanation), we measure recognition accuracy through: (\textit{a}) True Positive Rate (TPR) measures the proportion of correctly identified puns.
(\textit{b}) True Negative Rate (TNR) indicates the proportion of correctly identified non-puns. 
(\textit{c}) F1-Score provides an overall performance assessment.
(\textit{d}) Variations ($\Delta$) in TPR and TNR when the prompt leans towards non-pun compared to pun.
(\textit{e}) Cohen's Kappa ($\kappa$)~\cite{cohen1960coefficient} measures agreement between two sets of biased recognitions.
(\textit{ii}) \textbf{Word Extraction and Explanation Quality.} For localization and explanation tasks, we use: 
(\textit{a}) Mention ratio measures the proportion of ground-truth $w_p$ and $w_a$ in the extracted tuples that models correctly identify puns. 
(\textit{b}) Win/tie/loss rates measure the judge's result by comparing model-generated explanations to ground-truth explanations.

% Direct Detection
% Definition Prompting
% Structured Localization
% Reasoning Explanation
% (Direct)、(Def)、(Def+Loc)、(Def+Loc+Exp)

\definecolor{myred}{HTML}{F06060}
\definecolor{myyellow}{HTML}{F57A0C}
\definecolor{mycyan}{HTML}{007DEA}
\definecolor{mygreen}{HTML}{28A745}
\definecolor{mygreen}{HTML}{007369}

\def\myredscoremin{0.75}
\def\myredscoremax{1.4}
\def\mycyanscoremin{0}
\def\mycyanscoremax{1.5}
\def\mygreenscoremin{0.4}
\def\mygreenscoremax{1.2}
\def\myyellowscoremin{0}
\def\myyellowscoremax{1}

\newcommand{\pmin}{12}
\newcommand{\pmax}{75}

\newcommand{\colorcell}[2]{%
  % #1: 基色名（如 mycyan 或自定义颜色名）
  % #2: 数值（百分制，如 68）

  % —— 根据 #1 解析出 \scoremin / \scoremax —— 
  % 如果 \#1scoremin / \#1scoremax 存在就用之，否则回退到默认
  \ifcsname #1scoremin\endcsname
    \edef\scoremin{\csname #1scoremin\endcsname}%
  \else
    \edef\scoremin{\defaultscoremin}%
  \fi
  \ifcsname #1scoremax\endcsname
    \edef\scoremax{\csname #1scoremax\endcsname}%
  \else
    \edef\scoremax{\defaultscoremax}%
  \fi

  % \pgfmathsetmacro{\norm}{(#2/100 - \scoremin)/(\scoremax - \scoremin)}%
  \pgfmathsetmacro{\norm}{(#2 - \scoremin)/(\scoremax - \scoremin)}%
  \pgfmathsetmacro{\t}{max(0, min(1, \norm))}
  \pgfmathsetmacro{\praw}{\pmin + \t*(\pmax-\pmin)}%
  \pgfmathtruncatemacro{\pct}{\praw}

  % 用 \edef 冻结成字面量：此时 {#1!\pct} 里 \pct 已被展开成纯数字
  \edef\__tmp{\noexpand\cellcolor{#1!\pct}{#2}}%
  \__tmp%
}

\newif\ifopcolor
\opcolortrue      % 着色：\opcolorcell == \colorcell
% \opcolorfalse   % 关色：\opcolorcell 只输出 #2

\DeclareRobustCommand{\opcolorcell}[2]{%
  \ifopcolor
    \colorcell{#1}{#2}%
  \else
    #2%
  \fi
}

% =========================================================
% 2. 表格正文
% =========================================================
\begin{table*}[t]
    \centering
    \small
    \setlength{\aboverulesep}{1.5pt} 
    \setlength{\belowrulesep}{1.5pt}
    
    \resizebox{\linewidth}{!}{
    \setlength{\tabcolsep}{4pt}
    \begin{tabular}{c|c|c|>{\centering\arraybackslash}p{0.06\linewidth}c>{\centering\arraybackslash}p{0.06\linewidth}c>{\centering\arraybackslash}p{0.06\linewidth}>{\centering\arraybackslash}p{0.06\linewidth}|>{\centering\arraybackslash}p{0.06\linewidth}c>{\centering\arraybackslash}p{0.06\linewidth}c>{\centering\arraybackslash}p{0.06\linewidth}>{\centering\arraybackslash}p{0.06\linewidth}} \toprule
    \multirow{2}{*}{\textbf{Type}} & \multirow{2}{*}{\makecell[c]{\textbf{Model}}} & \multirow{2}{*}{\makecell[c]{\textbf{Task}}} & \multicolumn{6}{c|}{\textbf{Homophonic Pun}} & \multicolumn{6}{c}{\textbf{Homographic Pun}} \\
    \cmidrule(lr){4-9} \cmidrule(lr){10-15}
     & & & TPR $\uparrow$ & $\Delta$\textsubscript{TPR} $\downarrow$ & TNR $\uparrow$ & $\Delta$\textsubscript{TNR} $\downarrow$ & F1 $\uparrow$ & $\kappa$ $\uparrow$ & TPR $\uparrow$ & $\Delta$\textsubscript{TPR} $\downarrow$ & TNR $\uparrow$ & $\Delta$\textsubscript{TNR} $\downarrow$ & F1 $\uparrow$ & $\kappa$ $\uparrow$ \\ \midrule

     % ================= Group 1: Closed-Source VLMs =================
    \multirow{12}{*}{\rotatebox[origin=c]{90}{Closed-Source VLMs}}
    % ----- GPT-5.1 -----
     & \multirow{3}{*}{\makecell[c]{GPT-5.1}}
     & Detection    & \bf \opcolorcell{myred}{0.933} & \textbf{-0.026} & \colorcell{mycyan}{0.379} & +0.198 & \colorcell{mygreen}{0.588} & \colorcell{myyellow}{0.241} & \bf \opcolorcell{myred}{0.956} & \textbf{-0.036} & \colorcell{mycyan}{0.243} & +0.201 & \colorcell{mygreen}{0.551} & \colorcell{myyellow}{0.146} \\
     & & Localization & \colorcell{myred}{0.887} & -0.046 & \colorcell{mycyan}{0.768} & +0.072 & \colorcell{mygreen}{0.754} & \colorcell{myyellow}{0.601} & \colorcell{myred}{0.876} & -0.108 & \colorcell{mycyan}{0.695} & +0.141 & \colorcell{mygreen}{0.705} & \colorcell{myyellow}{0.508} \\
     & & Explanation  & \colorcell{myred}{0.794} & -0.062 & \bf \opcolorcell{mycyan}{0.910} & \textbf{+0.059} & \bf \opcolorcell{mygreen}{0.804} & \bf \opcolorcell{myyellow}{0.708} & \colorcell{myred}{0.757} & -0.143 & \bf \opcolorcell{mycyan}{0.878} & \textbf{+0.060} & \bf \opcolorcell{mygreen}{0.757} & \bf \opcolorcell{myyellow}{0.637} \\ \cmidrule{2-15}
    
    % ----- GPT-4o -----
     & \multirow{3}{*}{\makecell[c]{GPT-4o}}
     & Detection    & \bf \opcolorcell{myred}{0.933} & \textbf{0.000} & \colorcell{mycyan}{0.332} & +0.144 & \colorcell{mygreen}{0.571} & \colorcell{myyellow}{0.202} & \bf \opcolorcell{myred}{0.956} & \textbf{-0.004} & \colorcell{mycyan}{0.211} & +0.122 & \colorcell{mygreen}{0.541} & \colorcell{myyellow}{0.121} \\
     & & Localization & \colorcell{myred}{0.923} & -0.015 & \colorcell{mycyan}{0.582} & +0.088 & \colorcell{mygreen}{0.669} & \colorcell{myyellow}{0.425} & \colorcell{myred}{0.888} & -0.028 & \colorcell{mycyan}{0.480} & +0.120 & \colorcell{mygreen}{0.607} & \colorcell{myyellow}{0.299} \\
     & & Explanation  & \colorcell{myred}{0.840} & -0.026 & \bf \opcolorcell{mycyan}{0.786} & \textbf{+0.072} & \bf \opcolorcell{mygreen}{0.741} & \bf \opcolorcell{myyellow}{0.587} & \colorcell{myred}{0.873} & -0.064 & \bf \opcolorcell{mycyan}{0.659} & \textbf{+0.096} & \bf \opcolorcell{mygreen}{0.683} & \bf \opcolorcell{myyellow}{0.467} \\ \cmidrule{2-15}

    % ----- Gemini -----
     & \multirow{3}{*}{\makecell[c]{Gemini-3-Pro}}
     & Detection    & \bf \opcolorcell{myred}{0.979} & -0.015 & \colorcell{mycyan}{0.268} & +0.142 & \colorcell{mygreen}{0.569} & \colorcell{myyellow}{0.181} & \colorcell{myred}{0.984} & -0.008 & \colorcell{mycyan}{0.209} & +0.135 & \colorcell{mygreen}{0.552} & \colorcell{myyellow}{0.139} \\
     & & Localization & \colorcell{myred}{0.974} & +0.005 & \colorcell{mycyan}{0.250} & +0.039 & \colorcell{mygreen}{0.561} & \colorcell{myyellow}{0.163} & \bf \opcolorcell{myred}{0.996} & -0.016 & \colorcell{mycyan}{0.221} & +0.064 & \colorcell{mygreen}{0.561} & \colorcell{myyellow}{0.158} \\
     & & Explanation  & \colorcell{myred}{0.969} & \textbf{-0.005} & \bf \opcolorcell{mycyan}{0.686} & \textbf{+0.023} & \bf \opcolorcell{mygreen}{0.746} & \bf \opcolorcell{myyellow}{0.579} & \colorcell{myred}{0.980} & \textbf{-0.004} & \bf \opcolorcell{mycyan}{0.625} & \textbf{+0.008} & \bf \opcolorcell{mygreen}{0.718} & \bf \opcolorcell{myyellow}{0.520} \\ \cmidrule{2-15}

    % ----- Claude -----
     & \multirow{3}{*}{\makecell[c]{Claude\\Sonnet-4.5}}
      & Detection    & \colorcell{myred}{0.974} & \textbf{-0.005} & \colorcell{mycyan}{0.134} & +0.134 & \colorcell{mygreen}{0.526} & \colorcell{myyellow}{0.076} & \colorcell{myred}{0.992} & -0.012 & \colorcell{mycyan}{0.102} & +0.110 & \colorcell{mygreen}{0.524} & \colorcell{myyellow}{0.065} \\
      & & Localization & \bf \opcolorcell{myred}{0.990} & +0.010 & \colorcell{mycyan}{0.072} & +0.072 & \colorcell{mygreen}{0.515} & \colorcell{myyellow}{0.042} & \bf \opcolorcell{myred}{0.996} & \textbf{0.000} & \colorcell{mycyan}{0.046} & \textbf{+0.052} & \colorcell{mygreen}{0.510} & \colorcell{myyellow}{0.028} \\
      & & Explanation  & \colorcell{myred}{0.969} & -0.010 & \bf \opcolorcell{mycyan}{0.353} & \textbf{+0.070} & \bf \opcolorcell{mygreen}{0.594} & \bf \opcolorcell{myyellow}{0.245} & \colorcell{myred}{0.984} & +0.004 & \bf \opcolorcell{mycyan}{0.235} & +0.127 & \bf \opcolorcell{mygreen}{0.560} & \bf \opcolorcell{myyellow}{0.159} \\ \midrule

    % ================= Group 2: Open-Source VLMs =================
    \multirow{12}{*}{\rotatebox[origin=c]{90}{\makecell{Open-Source VLMs}}}
    % ----- Qwen3-VL-8B-Instruct -----
     & \multirow{3}{*}{\makecell[c]{Qwen3-VL\\8B-Instruct}}
      & Detection    & \bf \opcolorcell{myred}{0.923} & \textbf{-0.160} & \colorcell{mycyan}{0.193} & +0.338 & \colorcell{mygreen}{0.522} & \colorcell{myyellow}{0.084} & \bf \opcolorcell{myred}{0.968} & -0.263 & \colorcell{mycyan}{0.147} & +0.351 & \bf \opcolorcell{mygreen}{0.527} & \colorcell{myyellow}{0.082} \\
      & & Localization & \colorcell{myred}{0.799} & -0.222 & \colorcell{mycyan}{0.487} & \textbf{+0.291} & \bf \opcolorcell{mygreen}{0.566} & \colorcell{myyellow}{0.237} & \colorcell{myred}{0.681} & -0.359 & \colorcell{mycyan}{0.490} & +0.307 & \colorcell{mygreen}{0.504} & \bf \opcolorcell{myyellow}{0.146} \\
      & & Explanation  & \colorcell{myred}{0.418} & -0.268 & \bf \opcolorcell{mycyan}{0.881} & +0.111 & \colorcell{mygreen}{0.505} & \bf \opcolorcell{myyellow}{0.329} & \colorcell{myred}{0.207} & \textbf{-0.191} & \bf \opcolorcell{mycyan}{0.904} & \textbf{+0.084} & \colorcell{mygreen}{0.296} & \colorcell{myyellow}{0.131} \\ \cmidrule{2-15}

      % ----- Qwen3-VL-30B-A3B-Instruct -----
     & \multirow{3}{*}{\makecell[c]{Qwen3-VL\\30B-Instruct}}
      & Detection    & \bf \opcolorcell{myred}{0.990} & \textbf{-0.031} & \colorcell{mycyan}{0.018} & \textbf{+0.201} & \colorcell{mygreen}{0.501} & \colorcell{myyellow}{0.005} & \bf \opcolorcell{myred}{1.000} & \textbf{-0.048} & \colorcell{mycyan}{0.028} & \textbf{+0.506} & \colorcell{mygreen}{0.507} & \colorcell{myyellow}{0.019} \\
      & & Localization & \colorcell{myred}{0.985} & -0.129 & \colorcell{mycyan}{0.067} & +0.343 & \colorcell{mygreen}{0.511} & \colorcell{myyellow}{0.035} & \colorcell{myred}{0.996} & -0.155 & \colorcell{mycyan}{0.052} & +0.275 & \bf \opcolorcell{mygreen}{0.512} & \colorcell{myyellow}{0.033} \\
      & & Explanation  & \colorcell{myred}{0.943} & -0.273 & \bf \opcolorcell{mycyan}{0.209} & +0.469 & \bf \opcolorcell{mygreen}{0.535} & \bf \opcolorcell{myyellow}{0.110} & \colorcell{myred}{0.944} & -0.267 & \bf \opcolorcell{mycyan}{0.125} & +0.490 & \colorcell{mygreen}{0.511} & \bf \opcolorcell{myyellow}{0.050} \\ \cmidrule{2-15}

    % ----- LLaVA-v1.6-Vicuna-13B -----
     & \multirow{3}{*}{\makecell[c]{LLaVA-v1.6\\Vicuna-13B}}
      & Detection    & \bf \opcolorcell{myred}{0.969} & -0.923 & \colorcell{mycyan}{0.023} & +0.933 & \colorcell{mygreen}{0.494} & \colorcell{myyellow}{-0.005} & \bf \opcolorcell{myred}{0.980} & -0.944 & \colorcell{mycyan}{0.024} & +0.950 & \colorcell{mygreen}{0.498} & \colorcell{myyellow}{0.003} \\
      & & Localization & \colorcell{myred}{0.866} & -0.392 & \colorcell{mycyan}{0.072} & +0.356 & \bf \opcolorcell{mygreen}{0.465} & \colorcell{myyellow}{-0.043} & \colorcell{myred}{0.928} & -0.434 & \colorcell{mycyan}{0.102} & +0.359 & \colorcell{mygreen}{0.498} & \colorcell{myyellow}{0.021} \\
      & & Explanation  & \colorcell{myred}{0.031} & \textbf{-0.015} & \bf \opcolorcell{mycyan}{0.972} & \textbf{+0.023} & \colorcell{mygreen}{0.057} & \bf \opcolorcell{myyellow}{0.004} & \colorcell{myred}{0.028} & \textbf{-0.012} & \bf \opcolorcell{mycyan}{0.966} & \textbf{+0.026} & \colorcell{mygreen}{0.051} & \colorcell{myyellow}{-0.007} \\ \cmidrule{2-15}

    % ----- Llama-4-Scout-17B-Instruct -----
     & \multirow{3}{*}{\makecell[c]{Llama-4\\Scout-17B}}
      & Detection    & \colorcell{myred}{0.912} & -0.149 & \colorcell{mycyan}{0.423} & +0.381 & \colorcell{mygreen}{0.595} & \colorcell{myyellow}{0.265} & \bf \opcolorcell{myred}{0.912} & -0.275 & \colorcell{mycyan}{0.341} & +0.408 & \bf \opcolorcell{mygreen}{0.565} & \colorcell{myyellow}{0.193} \\
      & & Localization & \bf \opcolorcell{myred}{0.933} & \textbf{0.000} & \colorcell{mycyan}{0.407} & \textbf{-0.064} & \colorcell{mygreen}{0.598} & \colorcell{myyellow}{0.266} & \colorcell{myred}{0.837} & \textbf{+0.044} & \colorcell{mycyan}{0.355} & \textbf{-0.112} & \colorcell{mygreen}{0.535} & \colorcell{myyellow}{0.147} \\
      & & Explanation  & \colorcell{myred}{0.799} & -0.072 & \bf \opcolorcell{mycyan}{0.624} & +0.142 & \bf \opcolorcell{mygreen}{0.626} & \bf \opcolorcell{myyellow}{0.372} & \colorcell{myred}{0.749} & -0.100 & \bf \opcolorcell{mycyan}{0.494} & +0.145 & \colorcell{mygreen}{0.543} & \bf \opcolorcell{myyellow}{0.204} \\ \midrule

    % ================= Group 3: Open-Source Reasoning-based VLMs =================
% ================= Group 3: Open-Source Reasoning-based VLMs =================
    \multirow{9}{*}{\rotatebox[origin=c]{90}{\makecell{Open-Source\\Reasoning-based\\VLMs}}}

    % ----- GLM-4.1V-9B-Think -----
      & \multirow{3}{*}{\makecell[c]{GLM-4.1V\\9B-Thinking}}
      & Detection    & \bf \opcolorcell{myred}{0.969} & -0.206 & \colorcell{mycyan}{0.124} & +0.487 & \colorcell{mygreen}{0.521} & \colorcell{myyellow}{0.050} & \bf \opcolorcell{myred}{0.956} & -0.247 & \colorcell{mycyan}{0.092} & +0.484 & \colorcell{mygreen}{0.507} & \colorcell{myyellow}{0.026} \\
      & & Localization & \colorcell{myred}{0.887} & -0.129 & \colorcell{mycyan}{0.567} & +0.245 & \bf \opcolorcell{mygreen}{0.644} & \colorcell{myyellow}{0.367} & \colorcell{myred}{0.841} & -0.175 & \bf \opcolorcell{mycyan}{0.550} & +0.052 & \colorcell{mygreen}{0.613} & \colorcell{myyellow}{0.411} \\
      & & Explanation  & \colorcell{myred}{0.835} & \textbf{-0.015} & \bf \opcolorcell{mycyan}{0.629} & \textbf{+0.062} & \colorcell{mygreen}{0.648} & \bf \opcolorcell{myyellow}{0.376} & \colorcell{myred}{0.940} & \textbf{-0.044} & \colorcell{mycyan}{0.550} & \textbf{+0.052} & \bf \opcolorcell{mygreen}{0.662} & \bf \opcolorcell{myyellow}{0.411} \\ \cmidrule{2-15}

      % ----- Qwen3-VL-8B-Think -----
      & \multirow{3}{*}{\makecell[c]{Qwen3-VL\\8B-Thinking}}
      & Detection    & \bf \opcolorcell{myred}{0.990} & -0.211 & \colorcell{mycyan}{0.054} & +0.593 & \colorcell{mygreen}{0.510} & \colorcell{myyellow}{0.023} & \colorcell{myred}{0.980} & -0.215 & \colorcell{mycyan}{0.048} & +0.554 & \colorcell{mygreen}{0.505} & \colorcell{myyellow}{0.016} \\
      & & Localization & \colorcell{myred}{0.985} & \textbf{-0.031} & \colorcell{mycyan}{0.106} & +0.263 & \colorcell{mygreen}{0.522} & \colorcell{myyellow}{0.090} & \bf \opcolorcell{myred}{0.992} & -0.052 & \colorcell{mycyan}{0.118} & +0.309 & \colorcell{mygreen}{0.528} & \colorcell{myyellow}{0.117} \\
      & & Explanation  & \colorcell{myred}{0.943} & -0.077 & \bf \opcolorcell{mycyan}{0.387} & \textbf{+0.119} & \bf \opcolorcell{mygreen}{0.595} & \bf \opcolorcell{myyellow}{0.325} & \colorcell{myred}{0.960} & \textbf{-0.044} & \bf \opcolorcell{mycyan}{0.367} & \textbf{+0.197} & \bf \opcolorcell{mygreen}{0.595} & \bf \opcolorcell{myyellow}{0.343} \\ \cmidrule{2-15}

    % ----- Qwen3-VL-30B-A3B-Thinking -----
      & \multirow{3}{*}{\makecell[c]{Qwen3-VL\\30B-A3B\\Thinking}}
      & Detection    & \colorcell{myred}{0.990} & -0.149 & \colorcell{mycyan}{0.106} & +0.448 & \colorcell{mygreen}{0.524} & \colorcell{myyellow}{0.049} & \colorcell{myred}{0.992} & -0.112 & \colorcell{mycyan}{0.078} & +0.390 & \colorcell{mygreen}{0.517} & \colorcell{myyellow}{0.036} \\
      & & Localization & \bf \opcolorcell{myred}{1.000} & \textbf{0.000} & \colorcell{mycyan}{0.165} & +0.227 & \bf \opcolorcell{mygreen}{0.545} & \colorcell{myyellow}{0.145} & \bf \opcolorcell{myred}{1.000} & \textbf{-0.008} & \colorcell{mycyan}{0.151} & +0.319 & \bf \opcolorcell{mygreen}{0.541} & \colorcell{myyellow}{0.135} \\
      & & Explanation  & \colorcell{myred}{0.985} & -0.026 & \bf \opcolorcell{mycyan}{0.399} & \textbf{+0.155} & \colorcell{mygreen}{0.618} & \bf \opcolorcell{myyellow}{0.273} & \bf \opcolorcell{myred}{1.000} & -0.020 & \bf \opcolorcell{mycyan}{0.414} & \textbf{+0.163} & \colorcell{mygreen}{0.631} & \bf \opcolorcell{myyellow}{0.298} \\
    
    \bottomrule
    \end{tabular}
    }
    
    \vspace{-0.8em}
    \caption{Results of pun recognition in detection, localization, and explanation tasks. Metrics (TPR, TNR, F1, $\kappa$) are evaluated under the \textit{biased-to-pun} prompt. $\Delta$ measures variations when prompt bias shifts from pun to non-pun. Darker colors indicate better performance. The best results (smallest variations or highest scores) are \textbf{bolded}.}
    \vspace{-5px}
    \label{tab:model_comparison}
\end{table*}

% \vspace{-2px}
\section{Results and Analysis}
\label{experiments}
\vspace{-2px}
\subsection{How Effectively Can VLMs Recognize Multimodal Puns Against Non-puns?}
\label{sec:recognize}

Table~\ref{tab:model_comparison} shows the results of VLMs on pun recognition tasks, including detection, localization, and explanation. We have the following observations.

\noindent \textbf{VLMs often classify non-pun samples as puns.}
Most models achieve high TPR in pun recognition but struggle with low TNR, particularly in detection and localization tasks.
For example, Qwen3-VL-30B-A3B-Instruct identifies almost every input as a pun, achieving a near-perfect TPR of 0.990, but its TNR drops to 0.018 in detecting homophonic puns. 
Similarly, closed-source models such as GPT-5.1, GPT-4o, Gemini-3-Pro, and Claude-Sonnet-4.5 exhibit TNR scores mostly below 0.38 in detection tasks.
Even in the explanation task, although GPT-5.1 and GPT-4o improve their TNR to above 0.75, Gemini-3-Pro and Claude-Sonnet-4.5 remain lower at 0.686 and 0.353, respectively.
This imbalance results in poor Cohen's Kappa scores ($\kappa < 0.4$), indicating that models frequently misclassify non-puns as puns rather than a genuine understanding of pun mechanisms.

\noindent \textbf{Open-source models exhibit greater prompt-induced bias in pun recognition.} 
We measure prompt-induced bias (i.e., where model decisions are influenced by prompt phrasing rather than content) through the variations in $\Delta$\textsubscript{TPR} and $\Delta$\textsubscript{TNR} when switching from \textit{biased-to-pun} prompt to \textit{biased-to-non-pun} prompt. These variations reveal that many VLMs, particularly open-source ones, are easily influenced by the way questions are asked and lack robust internal reasoning for pun recognition. 
Notably, LLaVA-V1.6-Vicuna-13B exhibits a dramatic $\Delta$\textsubscript{TPR} of -0.923, suggesting that its decisions are primarily driven by prompt question format rather than the genuine multimodal understanding. 
In contrast, closed-source models such as GPT-4o and Gemini-3-Pro maintain consistency across prompt variations, with low absolute values of $\Delta$\textsubscript{TPR} and $\Delta$\textsubscript{TNR}, demonstrating superior robustness in reasoning.

\noindent \textbf{Explanation tasks improve non-pun rejection but slightly compromise pun detection.}
Models perform better at rejecting non-puns when tasked with explaining the pun rather than simply detecting or localizing it. A clear upward trend in TNR is observed across most models during explanation tasks. 
For instance,  the TNR of GPT-5.1 for homophonic puns increases sharply from 0.379 in detection to 0.910 in explanation. 
This suggests that requiring models to explicitly identify pun components and explain their reasoning helps ground their judgments in evidence, effectively reducing hallucinated false positives. However, this stricter verification process consistently leads to a drop in TPR, indicating that models sometimes discard valid puns when they fail to correctly explain the underlying punning mechanism.

\noindent \textbf{Closed-source models outperform open-source counterparts in pun recognition.} 
% Closed-source models such as GPT-5.1, GPT-4o, and Gemini-3-Pro show high F1 scores. In contrast, open-source models often struggle to recognize puns. A notable example is LLaVA-V1.6-Vicuna-13B, whose performance collapses in the explanation task, with the F1 score dropping to approximately 0.058. This failure suggests the deficiency in pun comprehension.
Closed-source models such as GPT-5.1, GPT-4o, and Gemini-3-Pro consistently demonstrate superior performance across detection, localization, and explanation tasks, achieving high F1 scores. In contrast, open-source models often struggle to recognize puns accurately, exhibiting lower F1 scores and more pronounced performance gaps between TPR and TNR. A notable example is LLaVA-V1.6-Vicuna-13B, whose performance collapses in the explanation task, with the F1 score dropping to approximately 0.058. This failure suggests deficiencies in pun comprehension, likely due to limited training data or architectural constraints.

\noindent \textbf{Reasoning-based models do not guarantee improved pun recognition.}
Comparing standard models with their reasoning-based ``Thinking'' variants reveals mixed results based on model scale. For smaller models such as Qwen3-VL-8B-Instruct, introducing reasoning processes worsens performance, with TNR dropping from 0.193 to 0.054, indicating hallucination in pun recognition. Conversely, larger models such as Qwen3-VL-30B-A3B-Instruct benefit from reasoning, improving both pun detection and non-pun rejection. Specifically, its TPR increases from 0.943 to 0.985, while its TNR improves from 0.209 to 0.399.

\noindent \textbf{Error analysis of negative samples.}
We categorize false positives into four distinct hallucination patterns, covering the lexical, phonological, semantic, and visual levels. 
(\textit{i}) \ul{\textit{Pun word hallucination.}} 
VLMs prioritize idiomatic priors over visual evidence. The model ignores the actual word written in the text and shown in the image (e.g., ``lamp'') and mistakenly imagines the common word that usually fits the idiom (e.g., ``fan''). 
(\textit{ii}) \ul{\textit{Phonetic hallucination.}}
To force a connection, the model wrongly claims that two words sound alike, even when they sound completely different (e.g., claiming ``banana'' sounds like ``soul'').
(\textit{iii}) \ul{\textit{Semantic hallucination.}}
Models correctly identify the alternative word $w_a$ but invent a meaning that does not exist. For instance, it tries to force the meaning of ``pair'' onto the word ``banana'', even though they are not related.
(\textit{iv}) \ul{\textit{Visual object hallucination.}}
Misled by the text, the model imagines seeing objects that are not actually in the image. For example, reading about a ``date'' makes the model say it sees the fruit ``date'' in the image, when it is actually an apple.
We provided detailed case studies in Appendix \ref{app:appendix_negative_cases}.

% (1) \textbf{Word semantic interpretation error.} 
% Models frequently hallucinate polysemy in non-pun captions to force a dual interpretation. This is particularly prevalent in Explicative Substitution, where the model encounters a non-ambiguous synonym (e.g., "supporter" instead of "fan"). In such cases, the model often incorrectly projects the visual object's polysemy onto the text, falsely asserting that the synonym shares the physical definition of the visual object (e.g., claiming "supporter" literally refers to an electric cooling device), conflating the semantic properties of the visual proxy with the actual textual token.
% % 词义联想（词义错误）：词语与含义之间的错误联想
% (2) \textbf{Phonetic/Spelling Hallucination.} For non-puns where the keyword is replaced by a semantically related but phonetically distinct word (e.g., Random Substitution), models frequently fabricate phonetic similarities under contextual bias. Driven by visual cues (e.g., seeing two apples implying a "pair") or the strong prior of the underlying idiom, the model incorrectly asserts that the substitute word acts as a homophone (e.g., claiming "apple" sounds like "pair") to justify a positive classification.
% % 双关语对：在语音/拼写方面不够相似，无法构成文字游戏

% \definecolor{mypurple}{HTML}{5477b0}
% \definecolor{mydarkpurple}{HTML}{8158a2}
% \definecolor{myyellow}{HTML}{F57A0C}
% \definecolor{mycyan}{HTML}{007DEA}
% \definecolor{mygreen}{HTML}{007369}
% % 原 e5f4fb 的加深版 (天空蓝)
% \definecolor{mypurple}{HTML}{B3E5FC} 

% % 原 fbe7e7 的加深版 (淡玫瑰红)
% \definecolor{mydarkpurple}{HTML}{FFCDD2}
\definecolor{mypurple}{HTML}{F06060} 
\definecolor{mydarkpurple}{HTML}{007DEA}
\def\mypurplescoremin{60.}
\def\mypurplescoremax{180.}
\def\mydarkpurplescoremin{80.}
\def\mydarkpurplescoremax{158.}

\begin{table}[t]
    \centering
    \small
    
    \resizebox{\linewidth}{!}{
    \setlength{\tabcolsep}{2pt} % 保持紧凑的列间距
    % 9列结构：Model(1) + Homophonic(4) + Homographic(4)
    \begin{tabular}{>{\centering\arraybackslash}p{0.45\linewidth}|>{\centering\arraybackslash}p{0.1\linewidth}>{\centering\arraybackslash}p{0.1\linewidth}|>{\centering\arraybackslash}p{0.1\linewidth}>{\centering\arraybackslash}p{0.1\linewidth}|>{\centering\arraybackslash}p{0.1\linewidth}>{\centering\arraybackslash}p{0.1\linewidth}|>{\centering\arraybackslash}p{0.1\linewidth}>{\centering\arraybackslash}p{0.1\linewidth}} \toprule
    \multirow{3}{*}{\textbf{Model}} & \multicolumn{4}{c|}{\textbf{Homophonic Pun}} & \multicolumn{4}{c}{\textbf{Homographic Pun}} \\
    \cmidrule(lr){2-5} \cmidrule(lr){6-9}
     & \multicolumn{2}{c|}{\footnotesize Localization} & \multicolumn{2}{c|}{\footnotesize Explanation} & \multicolumn{2}{c|}{\footnotesize Localization} & \multicolumn{2}{c}{\footnotesize Explanation} \\
    \cmidrule(lr){2-3} \cmidrule(lr){4-5} \cmidrule(lr){6-7} \cmidrule(lr){8-9}
     & \cellcolor{red!10}{$w_p$} & \cellcolor{cyan!10}{$w_a$} & \cellcolor{red!10}{$w_p$} & \cellcolor{cyan!10}{$w_a$} & \cellcolor{red!10}{$w_p$} & \cellcolor{cyan!10}{$w_a$} & \cellcolor{red!10}{$w_p$} & \cellcolor{cyan!10}{$w_a$} \\ \midrule

    % ================= Closed-Source VLMs =================
    \multicolumn{9}{c}{\textit{\textbf{Closed-Source VLMs}}} \\ \midrule
    GPT-5.1 
    & \bf \colorcell{mypurple}{98.8} & \bf \colorcell{mydarkpurple}{87.8} 
    & \bf \colorcell{mypurple}{100.0} & \bf \colorcell{mydarkpurple}{89.0} 
    & \colorcell{mypurple}{97.7} & \colorcell{mydarkpurple}{97.7} 
    & \colorcell{mypurple}{97.9} & \colorcell{mydarkpurple}{97.9} \\
    
    GPT-4o 
    & \colorcell{mypurple}{96.1} & \colorcell{mydarkpurple}{84.9} 
    & \colorcell{mypurple}{92.6} & \colorcell{mydarkpurple}{75.5} 
    & \colorcell{mypurple}{97.3} & \colorcell{mydarkpurple}{97.3} 
    & \colorcell{mypurple}{97.3} & \colorcell{mydarkpurple}{97.3} \\
    
    Gemini-3-Pro 
    & \colorcell{mypurple}{97.4} & \colorcell{mydarkpurple}{86.8} 
    & \colorcell{mypurple}{97.9} & \colorcell{mydarkpurple}{88.8} 
    & \bf \colorcell{mypurple}{98.8} & \bf \colorcell{mydarkpurple}{98.8} 
    & \bf \colorcell{mypurple}{98.8} & \bf \colorcell{mydarkpurple}{98.8} \\
    
    Claude-Sonnet-4.5 
    & \colorcell{mypurple}{93.2} & \colorcell{mydarkpurple}{82.8} 
    & \colorcell{mypurple}{94.7} & \colorcell{mydarkpurple}{81.9} 
    & \colorcell{mypurple}{96.8} & \colorcell{mydarkpurple}{96.8} 
    & \colorcell{mypurple}{96.8} & \colorcell{mydarkpurple}{96.8} \\ \midrule

    % ================= Open-Source VLMs =================
    \multicolumn{9}{c}{\textit{\textbf{Open-Source VLMs}}} \\ \midrule
    Qwen3-VL-8B-Instruct 
    & \bf \colorcell{mypurple}{92.3} & \colorcell{mydarkpurple}{73.5} 
    & \bf \colorcell{mypurple}{90.1} & \colorcell{mydarkpurple}{40.7} 
    & \bf \colorcell{mypurple}{96.5} & \bf \colorcell{mydarkpurple}{96.5} 
    & \bf \colorcell{mypurple}{96.2} & \bf \colorcell{mydarkpurple}{96.2} \\
    
    Qwen3-VL-30B-Instruct 
    & \colorcell{mypurple}{84.3} & \colorcell{mydarkpurple}{75.4} 
    & \colorcell{mypurple}{82.5} & \colorcell{mydarkpurple}{59.0} 
    & \colorcell{mypurple}{96.0} & \colorcell{mydarkpurple}{96.0} 
    & \colorcell{mypurple}{94.5} & \colorcell{mydarkpurple}{94.5} \\
    
    LLaVA-v1.6-Vicuna-13B 
    & \colorcell{mypurple}{79.2} & \colorcell{mydarkpurple}{38.7} 
    & \colorcell{mypurple}{50.0} & \bf \colorcell{mydarkpurple}{83.3} 
    & \colorcell{mypurple}{91.0} & \colorcell{mydarkpurple}{91.0} 
    & \colorcell{mypurple}{42.9} & \colorcell{mydarkpurple}{42.9} \\ 
    
    Llama-4-Scout-17B 
    & \colorcell{mypurple}{91.7} & \bf \colorcell{mydarkpurple}{84.0} 
    & \colorcell{mypurple}{81.9} & \colorcell{mydarkpurple}{29.7} 
    & \colorcell{mypurple}{91.9} & \colorcell{mydarkpurple}{91.9} 
    & \colorcell{mypurple}{93.6} & \colorcell{mydarkpurple}{93.6} \\ \midrule

    % ================= Reasoning-based VLMs =================
    \multicolumn{9}{c}{\textit{\textbf{Open-Source Reasoning-Based VLMs}}} \\ \midrule
    GLM-4.1V-9B-Thinking 
    & \colorcell{mypurple}{96.5} & \colorcell{mydarkpurple}{80.8} 
    & \colorcell{mypurple}{86.4} & \colorcell{mydarkpurple}{59.3} 
    & \colorcell{mypurple}{98.1} & \colorcell{mydarkpurple}{98.1} 
    & \colorcell{mypurple}{95.8} & \colorcell{mydarkpurple}{95.8} \\
    
    Qwen3-VL-8B-Thinking 
    & \colorcell{mypurple}{94.8} & \colorcell{mydarkpurple}{81.7} 
    & \bf \colorcell{mypurple}{95.6} & \colorcell{mydarkpurple}{68.3} 
    & \colorcell{mypurple}{96.8} & \colorcell{mydarkpurple}{96.8} 
    & \colorcell{mypurple}{97.9} & \colorcell{mydarkpurple}{97.9} \\
    
    Qwen3-VL-30B-Thinking 
    & \bf \colorcell{mypurple}{96.9} & \bf \colorcell{mydarkpurple}{90.7} 
    & \colorcell{mypurple}{94.2} & \bf \colorcell{mydarkpurple}{81.2} 
    & \bf \colorcell{mypurple}{100.0} & \bf \colorcell{mydarkpurple}{100.0} 
    & \bf \colorcell{mypurple}{98.4} & \bf \colorcell{mydarkpurple}{98.4} \\

    \bottomrule
    \end{tabular}
    }
    
    \vspace{-0.5em}
    \caption{Pun component verification for pun localization and explanation. We represent the average mention ratio of the pun words $w_p$ and alternative words $w_a$.}
    \vspace{-10px}
    \label{tab:ppa_results}
\end{table}

\subsection{\!To What Extent Can VLMs Explain Puns?}
Beyond recognition, we explore pun understanding by: (\textit{i}) \textbf{pun component verification} check how accurately pun words $w_p$ and alternatives $w_a$ are identified, and (\textit{ii}) \textbf{explanation pairwise comparison} assesses the quality of the pun explanation.

\subsubsection{Pun Component Verification}
We calculate mention ratios for verifying the pun word $w_p$ and the alternative word $w_a$. As shown in Table~\ref{tab:ppa_results}, we have the following observations.

\noindent \textbf{VLMs accurately identify the pun word $w_p$.}
The mention ratio of $w_p$ remains consistently high across most models for both homophonic and homographic puns. 
For example, closed-source models such as GPT-5.1 and reasoning-based models such as Qwen3-VL-30B-A3B-Thinking achieve mention ratios over 94\%. 
Even smaller open-source models perform well (e.g., Qwen3-VL-8B-Instruct achieves 92.3\% in homophonic pun localization).
This high accuracy is due to $w_p$ appearing directly in the caption, making it easy to identify.

\noindent \textbf{Identifying the alternative word $w_a$ is the bottleneck for homophonic puns.}
Comparing the mention ratio of $w_p$, we observe a decrease in the mention ratio of $w_a$.
For instance, while Qwen3-VL-8B-Instruct achieves a 90.1\% mention ratio for $w_p$ in the explanation task, its performance on $w_a$ drops drastically to 40.7\%. 
This challenge arises because $w_a$ in homophonic puns does not directly appear in the text but depends on semantic inference and similar pronunciation to $w_p$. 

\noindent \textbf{Reasoning improves pun component identification.}
Compared to instruction-based models, reasoning-based models show a superior ability to identify both $w_p$ and $w_a$ through explicit thinking steps.
For example, for homophonic puns, Qwen3-VL-30B-A3B-Thinking increases the mention ratio of $w_a$ from 59.0\% (Instruct version) to 81.2\% in the explanation task. It also achieves highest mention ratio of both $w_p$ and $w_a$ on homographic puns (100\% in the localization task and 98.4\% in the explanation task). 
This suggests that the extended reasoning phase helps the model to explore phonetic or semantic connections more effectively.

\subsubsection{Explanation Pairwise Comparison}
\label{sec:explanation_pairwise_comparison}
\begin{figure}[t]
\begin{center}
\centerline{\includegraphics[width=\linewidth]{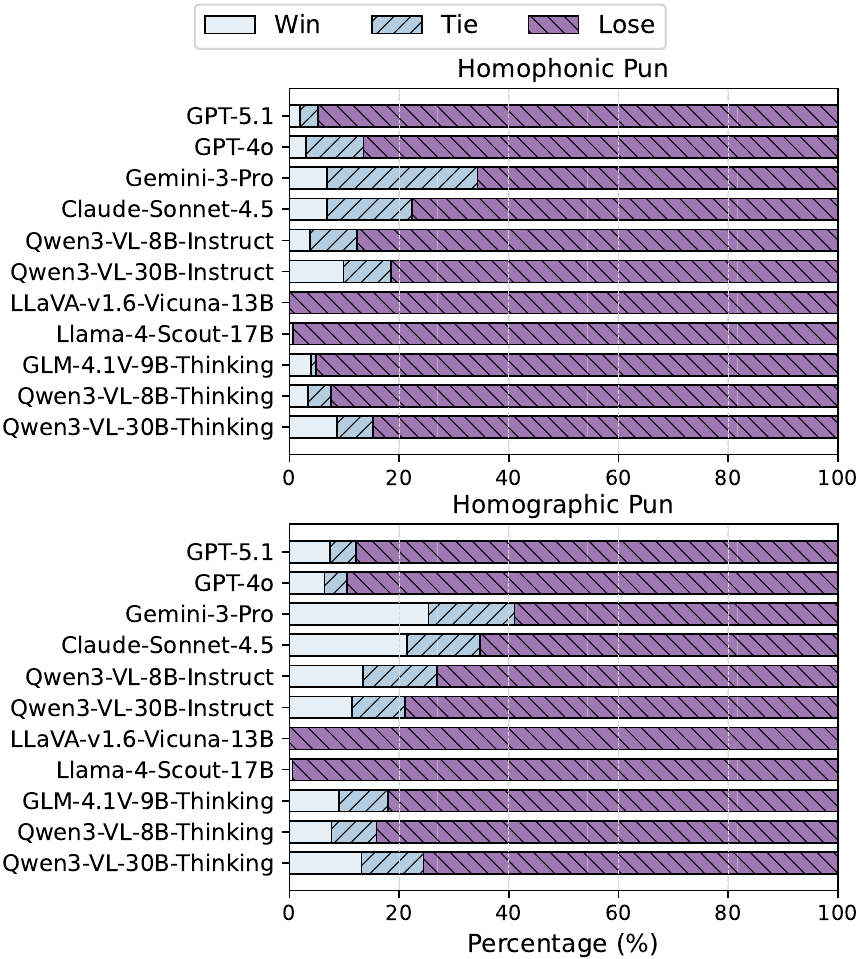}}
\vspace{-5px}
\caption{Pairwise comparison for pun explanations.} 
\label{fig:pairwise}
\end{center}
\vspace{-25px}
\end{figure}

While pun component verification measures recall on pun words, it does not assess the quality of the pun explanation. To evaluate this, we conduct a pairwise comparison where an advanced LLM judge compares the VLM-generated explanation to the ground-truth explanation from the \textsc{MultiPun} dataset. The judge classifies the comparison as a \textit{Win} (VLM is better), \textit{Tie} (Comparable), or \textit{Loss} (Ground truth is better).  As shown in Figure~\ref{fig:pairwise}, we have the following observations.

\noindent \textbf{Ground-truth explanations generally outperform VLM-generated explanations.} 
Across all evaluated models, the loss rate is exceptionally higher than the win rate. For instance, even the advanced GPT-5.1 loses to the ground truth in about 90\% of cases. This suggests that while models can identify pun components, recognizing them does not necessarily mean they understand the underlying logic of the pun effectively.

\noindent \textbf{VLMs explain homographic puns better than homophonic puns.}
We observe a consistent trend where models achieve higher win rates on homographic puns compared to homophonic ones. 
This aligns with the findings from the pun component verification and findings by ~\citet{xu-etal-2024-good}, where VLMs are better at explaining a word's multiple meanings than at articulating phonetic bridges by finding an alternative word $w_a$. 
Thus, alternative words do not affect pun recognition but are crucial for explaining puns more effectively.

\subsubsection{Error Analysis in Pun Explanation}
VLMs exhibit distinct error patterns in explaining puns. We categorize the primary errors as follows:
(\textit{i}) \ul{\textit{Detection Failure.}} VLMs identify pun as non-pun, failing to recognize the double meaning.
(\textit{ii}) \ul{\textit{Pun Word Error.}} VLMs detect the pun but fails to identify the pun word $w_p$. 
(\textit{iii}) \ul{\textit{Alternative Word Error.}} VLMs identify the correct pun word $w_p$ but fails to retrieve the intended alternative word $w_a$.
(\textit{iv}) \ul{\textit{Cross-modal Integration Error.}} VLMs identify both visual and textual content but explain them separately, failing to integrate them with the proper linguistic mechanism. 
We provide cases for each error type in Appendix \ref{app:appendix_positive_cases}. We believe that addressing these errors is pivotal to advancing VLMs' capability to recognize and understand puns.

\subsection{How Can We Enhance VLMs' Understanding of Puns?}
\begin{table}[t]
    \centering
    \small
    \setlength{\aboverulesep}{1.5pt} 
    \setlength{\belowrulesep}{1.5pt}
    
    % 调整宽度以适应页面
    \resizebox{\linewidth}{!}{
    \setlength{\tabcolsep}{3pt}
    \begin{tabular}{c|c|>{\centering\arraybackslash}p{0.115\linewidth}>{\centering\arraybackslash}p{0.115\linewidth}>{\centering\arraybackslash}p{0.115\linewidth}|>{\centering\arraybackslash}p{0.115\linewidth}>{\centering\arraybackslash}p{0.115\linewidth}>{\centering\arraybackslash}p{0.115\linewidth}} \toprule
    \multirow{2}{*}{\textbf{Model}} & \multirow{2}{*}{\textbf{Method}} & \multicolumn{3}{c|}{\textbf{Homophonic Pun}} & \multicolumn{3}{c}{\textbf{Homographic Pun}} \\
    \cmidrule(lr){3-5} \cmidrule(lr){6-8}
     & & TPR $\uparrow$ & TNR $\uparrow$ & F1 $\uparrow$ & TPR $\uparrow$ & TNR $\uparrow$ & F1 $\uparrow$ \\ \midrule

    % ================= GPT-5.1 Block =================
    \multirow{2}{*}{\makecell[c]{GPT-5.1}}
     &  Vanilla & \colorcell{myred}{0.794} & \colorcell{mycyan}{0.910} & \colorcell{mygreen}{0.804} & \colorcell{myred}{0.757} & \colorcell{mycyan}{0.878} & \colorcell{mygreen}{0.757} \\
     & \bf \cellcolor[HTML]{e4eef7}{Pun-CoT} & \bf \opcolorcell{myred}{0.840} & \bf \opcolorcell{mycyan}{0.915} & \bf \opcolorcell{mygreen}{0.836} & \bf \opcolorcell{myred}{0.813} & \bf \opcolorcell{mycyan}{0.894} & \bf \opcolorcell{mygreen}{0.803} \\ \midrule

    % ================= GPT-4o Block =================
    \multirow{2}{*}{\makecell[c]{GPT-4o}}
     &  Vanilla & \colorcell{myred}{0.840} & \colorcell{mycyan}{0.786} & \colorcell{mygreen}{0.741} & \colorcell{myred}{0.873} & \colorcell{mycyan}{0.659} & \colorcell{mygreen}{0.683} \\
     &  \bf \cellcolor[HTML]{e4eef7}{Pun-CoT} & \bf \opcolorcell{myred}{0.876} & \bf \opcolorcell{mycyan}{0.835} & \bf \opcolorcell{mygreen}{0.794} & \bf \opcolorcell{myred}{0.888} & \bf \opcolorcell{mycyan}{0.727} & \bf \opcolorcell{mygreen}{0.730} \\ \midrule

    % ================= Gemini Block =================
    \multirow{2}{*}{\makecell[c]{Gemini-3\\Pro}}
     &  Vanilla & \bf \opcolorcell{myred}{0.969} & \colorcell{mycyan}{0.686} & \colorcell{mygreen}{0.746} & \bf \opcolorcell{myred}{0.980} & \colorcell{mycyan}{0.625} & \colorcell{mygreen}{0.718} \\
     &  \bf \cellcolor[HTML]{e4eef7}{Pun-CoT} & \colorcell{myred}{0.959} & \bf \opcolorcell{mycyan}{0.719} & \bf \opcolorcell{mygreen}{0.761} & \colorcell{myred}{0.976} & \bf \opcolorcell{mycyan}{0.655} & \bf \opcolorcell{mygreen}{0.732} \\ \midrule

    % ================= Claude Block =================
    \multirow{2}{*}{\makecell[c]{Claude\\Sonnet-4.5}}
      &  Vanilla & \bf \opcolorcell{myred}{0.969} & \colorcell{mycyan}{0.353} & \colorcell{mygreen}{0.594} & \bf \opcolorcell{myred}{0.984} & \colorcell{mycyan}{0.235} & \colorcell{mygreen}{0.560} \\
      &  \bf \cellcolor[HTML]{e4eef7}{Pun-CoT} & \colorcell{myred}{0.948} & \bf \opcolorcell{mycyan}{0.495} & \bf \opcolorcell{mygreen}{0.641} & \colorcell{myred}{0.972} & \bf \opcolorcell{mycyan}{0.480} & \bf \opcolorcell{mygreen}{0.646} \\ \midrule

    % ================= Qwen3-VL-8B-Instruct Block =================
    \multirow{2}{*}{\makecell[c]{Qwen3-VL\\8B-Instruct}}
      &  Vanilla & \colorcell{myred}{0.418} & \bf \opcolorcell{mycyan}{0.881} & \colorcell{mygreen}{0.505} & \colorcell{myred}{0.207} & \bf \opcolorcell{mycyan}{0.904} & \colorcell{mygreen}{0.296} \\
      &  \bf \cellcolor[HTML]{e4eef7}{Pun-CoT} & \bf \opcolorcell{myred}{0.799} & \colorcell{mycyan}{0.495} & \bf \opcolorcell{mygreen}{0.569} & \bf \opcolorcell{myred}{0.685} & \colorcell{mycyan}{0.490} & \bf \opcolorcell{mygreen}{0.507} \\ \midrule
      
    % ================= Qwen3-VL-30B-Instruct Block =================
    \multirow{2}{*}{\makecell[c]{Qwen3-VL\\30B-Instruct}}
      &  Vanilla & \colorcell{myred}{0.943} & \colorcell{mycyan}{0.209} & \colorcell{mygreen}{0.535} & \colorcell{myred}{0.944} & \colorcell{mycyan}{0.125} & \colorcell{mygreen}{0.511} \\
      &  \bf \cellcolor[HTML]{e4eef7}{Pun-CoT} & \bf \opcolorcell{myred}{0.974} & \bf \opcolorcell{mycyan}{0.214} & \bf \opcolorcell{mygreen}{0.549} & \bf \opcolorcell{myred}{0.992} & \bf \opcolorcell{mycyan}{0.139} & \bf \opcolorcell{mygreen}{0.534} \\ \midrule

    % ================= LLaVA-v1.6-Vicuna-13B Block =================
    \multirow{2}{*}{\makecell[c]{LLaVA-v1.6\\Vicuna-13B}}
      &  Vanilla & \colorcell{myred}{0.031} & \bf \opcolorcell{mycyan}{0.972} & \colorcell{mygreen}{0.057} & \colorcell{myred}{0.028} & \bf \opcolorcell{mycyan}{0.966} & \colorcell{mygreen}{0.051} \\
      &  \bf \cellcolor[HTML]{e4eef7}{Pun-CoT} & \bf \opcolorcell{myred}{0.979} & \colorcell{mycyan}{0.036} & \bf \opcolorcell{mygreen}{0.501} & \bf \opcolorcell{myred}{0.984} & \colorcell{mycyan}{0.102} & \bf \opcolorcell{mygreen}{0.521} \\ \midrule

    % ================= Llama-4-Scout-17B-Instruct Block =================
    \multirow{2}{*}{\makecell[c]{Llama-4\\Scout-17B}}
      &  Vanilla & \colorcell{myred}{0.799} & \colorcell{mycyan}{0.624} & \colorcell{mygreen}{0.626} & \colorcell{myred}{0.749} & \colorcell{mycyan}{0.494} & \colorcell{mygreen}{0.543} \\
      &  \bf \cellcolor[HTML]{e4eef7}{Pun-CoT} & \bf \opcolorcell{myred}{0.866} & \bf \opcolorcell{mycyan}{0.629} & \bf \opcolorcell{mygreen}{0.664} & \bf \opcolorcell{myred}{0.757} & \bf \opcolorcell{mycyan}{0.522} & \bf \opcolorcell{mygreen}{0.558} \\ \midrule

    % ================= GLM-4.1V-9B-Thinking Block =================
    \multirow{2}{*}{\makecell[c]{GLM-4.1V\\9B-Thinking}}
      &  Vanilla & \colorcell{myred}{0.835} & \bf \opcolorcell{mycyan}{0.629} & \colorcell{mygreen}{0.648} & \bf \opcolorcell{myred}{0.940} & \colorcell{mycyan}{0.550} & \colorcell{mygreen}{0.662} \\
      &  \bf \cellcolor[HTML]{e4eef7}{Pun-CoT} & \bf \opcolorcell{myred}{0.948} & \colorcell{mycyan}{0.608} & \bf \opcolorcell{mygreen}{0.694} & \colorcell{myred}{0.916} & \bf \opcolorcell{mycyan}{0.757} & \bf \opcolorcell{mygreen}{0.763} \\ \midrule

    % ================= Qwen3-VL-8B-Thinking Block =================
    \multirow{2}{*}{\makecell[c]{Qwen3-VL\\8B-Thinking}}
      &  Vanilla & \colorcell{myred}{0.943} & \colorcell{mycyan}{0.387} & \colorcell{mygreen}{0.595} & \bf \opcolorcell{myred}{0.960} & \colorcell{mycyan}{0.367} & \colorcell{mygreen}{0.595} \\
      &  \bf \cellcolor[HTML]{e4eef7}{Pun-CoT} & \bf \opcolorcell{myred}{0.979} & \bf \opcolorcell{mycyan}{0.776} & \bf \opcolorcell{mygreen}{0.807} & \colorcell{myred}{0.920} & \bf \opcolorcell{mycyan}{0.797} & \bf \opcolorcell{mygreen}{0.791} \\ \midrule

    % ================= Qwen3-VL-30B-Thinking Block =================
    \multirow{2}{*}{\makecell[c]{Qwen3-VL\\30B-Thinking}}
      &  Vanilla & \bf \opcolorcell{myred}{0.985} & \colorcell{mycyan}{0.399} & \colorcell{mygreen}{0.618} & \bf \opcolorcell{myred}{1.000} & \colorcell{mycyan}{0.414} & \colorcell{mygreen}{0.631} \\
      &  \bf \cellcolor[HTML]{e4eef7}{Pun-CoT} & \colorcell{myred}{0.887} & \bf \opcolorcell{mycyan}{0.567} & \bf \opcolorcell{mygreen}{0.644} & \colorcell{myred}{0.976} & \bf \opcolorcell{mycyan}{0.480} & \bf \opcolorcell{mygreen}{0.647} \\

    \bottomrule
    \end{tabular}
    }
    
    \vspace{-0.5em}
    \caption{Comparison of pun recognition with and without Pun-CoT across VLMs under the explanation task.}
    \label{tab:defense_comparison}
    \vspace{-10px}
\end{table}

\subsubsection{Pun-CoT}
To mitigate the hallucinations identified in our error analysis, we propose \textbf{Pun-CoT}. 
Pun-CoT enforces the following process (see Appendix~\ref{appendix:pun_cot} for the complete prompt):
(\textit{i}) Visual Grounding. The model verifies the literal visual content to prevent visual object hallucinations.
(\textit{ii}) Lexical Anchoring. The model extracts exact keywords from the caption as $w_p$, thereby preventing hallucinated words not present in the caption.
(\textit{iii}) Cross-Modal Verification. The model checks if the visual content links to the text via a valid phonetic (for homophonic puns) or semantic (for homographic puns) bridge, rejecting weak or fabricated associations.

\noindent \textbf{Results.} Table~\ref{tab:defense_comparison} demonstrates the efficacy of Pun-CoT in balancing pun sensitivity with hallucination mitigation. Pun-CoT yields consistent improvements in F1 scores across diverse architectures, primarily driven by a substantial boost in TNR. Notably, for models prone to over-interpretation such as Qwen3-VL-8B-Thinking and Claude-Sonnet-4.5, Pun-CoT significantly enhances their ability to reject non-puns (e.g., doubling Qwen3-VL-8B-Thinking's TNR from 0.387 to 0.776) while maintaining competitive TPR. This confirms that explicitly grounding reasoning in verified visual and lexical evidence effectively filters out forced associations for robust comprehension.

\subsubsection{Pun-Tuning}

\noindent \textbf{Motivation.}
As illustrated in Section~\ref{sec:recognize}, current VLMs exhibit three challenges in pun understanding, including:
(\textit{i}) Over-interpretation, where models misclassify non-puns as puns due to a reliance on superficial pun pattern matching rather than a robust understanding;
(\textit{ii}) Imprecise explanations, revealing deficits in understanding fine-grained phonetic and orthographic similarity;
and (\textit{iii}) Prompt sensitivity, driven by alignment-induced \textit{sycophancy}, where models prioritize agreeableness with the user's premise over factual accuracy. 

To address these, our data construction includes: 
(\textit{i}) We incorporate non-pun samples to suppress hallucinations. 
(\textit{ii}) We utilize pun samples with high-quality responses to enhance recall and explanatory depth. 
(\textit{iii}) We employ both \textit{biased-to-pun} prompt and \textit{biased-to-non-pun} prompt. This improves robustness against prompt-induced bias.
We use the constructed dataset to fine-tune VLMs.
The implementation details are provided in Appendix~\ref{app:pun_tuning_details}.

\noindent \textbf{Results.}
Table~\ref{tab:tuning_comparison_detailed} reveals two key findings: 
(\textit{i}) Fine-tuning VLMs on non-pun samples enhances the non-pun recognition capabilities of fine-tuned models, as evidenced by improvements in the TNR and F1 scores. 
(\textit{ii}) Fine-tuning VLMs on pun samples enhances robustness against prompt-induced bias, with a decrease in the absolute values of $\Delta$\textsubscript{TPR} and $\Delta$\textsubscript{TNR}.
Additionally, we conduct the explanation pairwise comparison in the same way as Section~\ref{sec:explanation_pairwise_comparison}. As shown in Appendix~\ref{app:additional_results}, we observe that fine-tuning VLMs on pun samples enhances models' understanding of puns with a higher win rate compared to the model before fine-tuning.

\begin{table}[t]
    \centering
    \small
    \setlength{\aboverulesep}{1.5pt} 
    \setlength{\belowrulesep}{1.5pt}
    
    % 调整宽度以适应页面
    \resizebox{\linewidth}{!}{
    \setlength{\tabcolsep}{3pt}
    % 定义列格式
    \begin{tabular}{>{\centering\arraybackslash}p{0.06\linewidth}|c|c|>{\centering\arraybackslash}p{0.125\linewidth}>{\centering\arraybackslash}p{0.125\linewidth}>{\centering\arraybackslash}p{0.125\linewidth}>{\centering\arraybackslash}p{0.125\linewidth}>{\centering\arraybackslash}p{0.125\linewidth}} \toprule
    
    % --- 单行表头 ---
    & \textbf{Model} & \textbf{Method} & \textbf{TPR} $\uparrow$ & $\Delta$\textsubscript{TPR} $\downarrow$ & \textbf{TNR} $\uparrow$ & $\Delta$\textsubscript{TNR} $\downarrow$ & \textbf{F1} $\uparrow$ \\ \midrule

    % ================= BLOCK 1: Homophonic Pun =================
    \multirow{9}{*}{\rotatebox[origin=c]{90}{\textbf{Homophonic Pun}}} 
    
    % --- Qwen3-VL-8B ---
      & \multirow{2}{*}{\makecell[c]{Qwen3-VL\\8B-Instruct}} 
        & Vanilla & \colorcell{myred}{0.418} & -0.268 & \colorcell{mycyan}{0.881} & +0.111 & \colorcell{mygreen}{0.505} \\
      & & \bf \cellcolor[HTML]{cedef0}{Pun-Tuning} & \bf \opcolorcell{myred}{0.577} & \bf -0.155 & \bf \opcolorcell{mycyan}{0.938} & \bf +0.098 & \bf \opcolorcell{mygreen}{0.679} \\
      \cmidrule(lr){2-8}

    % --- Qwen3-VL-30B ---
      & \multirow{2}{*}{\makecell[c]{Qwen3-VL\\30B-Instruct}}
        & Vanilla & \bf \opcolorcell{myred}{0.943} & -0.273 & \colorcell{mycyan}{0.209} & +0.469 & \colorcell{mygreen}{0.535} \\
      & & \bf \cellcolor[HTML]{cedef0}{Pun-Tuning} & \colorcell{myred}{0.732} & \bf -0.062 & \bf \opcolorcell{mycyan}{0.948} & \bf +0.196 & \bf \opcolorcell{mygreen}{0.798} \\
      \cmidrule(lr){2-8}
      
    % --- LLaVA-v1.6 ---
      & \multirow{2}{*}{\makecell[c]{LLaVA-v1.6\\Vicuna-13B}}
        & Vanilla & \colorcell{myred}{0.031} & \bf -0.015 & \colorcell{mycyan}{0.972} & \bf +0.023 & \colorcell{mygreen}{0.057} \\
        & & \bf \cellcolor[HTML]{cedef0}{Pun-Tuning} & \bf \opcolorcell{myred}{0.495} & -0.103 & \bf \opcolorcell{mycyan}{0.974} & +0.098 & \bf \opcolorcell{mygreen}{0.640} \\
        \cmidrule(lr){2-8}

    % --- Llama-4-Scout ---
      & \multirow{2}{*}{\makecell[c]{Llama-4\\Scout-17B}}
        & Vanilla & \bf \opcolorcell{myred}{0.799} & \bf -0.072 & \colorcell{mycyan}{0.624} & +0.142 & \colorcell{mygreen}{0.626} \\
        & & \bf \cellcolor[HTML]{cedef0}{Pun-Tuning} & \colorcell{myred}{0.722} & -0.093 & \bf \opcolorcell{mycyan}{0.918} & \bf +0.119 & \bf \opcolorcell{mygreen}{0.765} \\ 
        \midrule

    % ================= BLOCK 2: Homographic Pun =================
    \multirow{9}{*}{\rotatebox[origin=c]{90}{\textbf{Homographic Pun}}} 
    
    % --- Qwen3-VL-8B ---
      & \multirow{2}{*}{\makecell[c]{Qwen3-VL\\8B-Instruct}} 
        & Vanilla & \colorcell{myred}{0.207} & -0.191 & \colorcell{mycyan}{0.904} & \bf +0.084 & \colorcell{mygreen}{0.296} \\
     & & \bf \cellcolor[HTML]{cedef0}{Pun-Tuning} & \bf \opcolorcell{myred}{0.556} & \bf -0.159 & \bf \opcolorcell{mycyan}{0.948} & +0.119 & \bf \opcolorcell{mygreen}{0.670} \\ \cmidrule(lr){2-8}

      % --- Qwen3-VL-30B ---
      & \multirow{2}{*}{\makecell[c]{Qwen3-VL\\30B-Instruct}}
        & Vanilla & \bf \opcolorcell{myred}{0.944} & \bf -0.267 & \colorcell{mycyan}{0.125} & +0.490 & \colorcell{mygreen}{0.511} \\
       & & \bf \cellcolor[HTML]{cedef0}{Pun-Tuning} & \colorcell{myred}{0.722} & -0.548 & \bf \opcolorcell{mycyan}{0.960} & \bf +0.222 & \bf \opcolorcell{mygreen}{0.802} \\
      \cmidrule(lr){2-8}

    % --- LLaVA-v1.6 ---
      & \multirow{2}{*}{\makecell[c]{LLaVA-v1.6\\Vicuna-13B}}
        & Vanilla & \colorcell{myred}{0.028} & \bf -0.012 & \colorcell{mycyan}{0.966} & \bf +0.026 & \colorcell{mygreen}{0.051} \\
        & & \bf \cellcolor[HTML]{cedef0}{Pun-Tuning} & \bf \opcolorcell{myred}{0.460} & -0.238 & \bf \opcolorcell{mycyan}{0.984} & +0.365 & \bf \opcolorcell{mygreen}{0.617} \\ \cmidrule(lr){2-8}

    % --- Llama-4-Scout ---
      & \multirow{2}{*}{\makecell[c]{Llama-4\\Scout-17B}}
        & Vanilla & \bf \opcolorcell{myred}{0.749} & \bf -0.100 & \colorcell{mycyan}{0.494} & +0.145 & \colorcell{mygreen}{0.543} \\
      & & \bf \cellcolor[HTML]{cedef0}{Pun-Tuning} & \colorcell{myred}{0.706} & -0.105 & \bf \opcolorcell{mycyan}{0.921} & \bf +0.103 & \bf \opcolorcell{mygreen}{0.757} \\

    \bottomrule
    \end{tabular}
    }
    
    \vspace{-0.5em}
    \caption{Comparison of pun recognition with and without Pun-Tuning on VLMs under the explanation task.}
    \vspace{-10px}

    \label{tab:tuning_comparison_detailed}
\end{table}

% \section{Discussion}
% \label{discussion}
% % \vspace{-1px}
% \input{5_discussions}

% \vspace{-2px}
\section{Conclusion}
\label{conclusion}
% \vspace{-1px}
In this paper, we propose \textsc{MultiPun}, a benchmark for evaluating VLMs' understanding of multimodal puns. Our benchmark includes both puns and non-puns. Through systematic evaluation of 11 VLMs across three pun recognition tasks--pun detection, localization, and explanation, we observe significant biases in pun recognition and deficits in understanding fine-grained phonetic and orthographic similarity of puns.
To enhance pun comprehension, we propose a prompt-level method, Pun-CoT, and a model-level method, Pun-Tuning. Our experiments show that both strategies improve VLMs' understanding of puns while preventing non-puns from being misidentified as puns.
We hope that our findings and the \textsc{MultiPun} benchmark will contribute to the advancement of multimodal pun understanding and encourage the development of more resilient and reliable VLM capabilities.

% \newpage
\section*{Limitations}
While \textsc{MultiPun} represents a significant step toward rigorous evaluation of multimodal pun comprehension, several limitations exist. 
First, our benchmark focuses exclusively on English puns. Since puns are deeply rooted in language-specific phonology, extending the dataset to other languages would test models' ability to handle multilingual settings. 
Second, our evaluation includes 11 representative VLMs spanning closed-source and open-source architectures, but newer models may exhibit different behaviors. 
Additionally, our fine-tuning experiments are limited to three open-source models due to computational constraints. Expanding fine-tuning experiments to more models and larger scales would strengthen our conclusions~\cite{xu2025fingerprintvectorenablingscalable}. Third, while our adversarial negatives effectively disrupt pun mechanisms, they may not cover all possible failure modes. Future work could design more diverse types of negative samples to probe model robustness comprehensively.

\section*{Ethics Considerations}

All data in \textsc{MultiPun} is generated using publicly available text-to-image models and language models, strictly following their intended purposes and respective licenses~\cite{xu2026bridging,xu2024copyrightmeter}. 
No personally identifiable information or real individuals are depicted in the images. All human annotators were compensated at rates exceeding local minimum wage standards and provided informed consent. 
The annotation task did not involve exposure to offensive, harmful, or distressing content. While advancements in pun understanding can enhance human-AI interaction~\cite{xu2026adamarpadaptivemultiagentinteraction,xu2026when}, we acknowledge the dual-use nature of such technologies, where AI systems capable of linguistic manipulation could be weaponized for social engineering or propaganda~\cite{xu2025copyrightprotectionlargelanguage,xuCTCCRobustStealthy2025}. We advocate for transparent reporting of model capabilities and limitations, as well as ongoing dialogue between researchers, ethicists, and policymakers to ensure responsible development and deployment~\cite{an-etal-2025-ipiguard,xu2025videoeraser,attardo2024linguistic}.

\section*{Acknowledgments}
This work was partly supported by the NSFC-Yeqisun Science Foundation under No. U244120033, NSFC under No. 62402418, Zhejiang Province's 2026 “Leading Goose + X” Science and Technology Plan under grant 2026C02A1233, the China Postdoctoral Science Foundation under No. 2024M762829, the Key R\&D Program of Ningbo under No. 2024Z115, and the Ningbo Yongjiang Talent Project.

\bibliography{custom}


\clearpage
\appendix
% \section*{Appendix}

\section{Dataset Statistics}
\label{appendix:statistics}

As shown in Table~\ref{tab:dataset_stats}, \textsc{MultiPun} comprises a total of 445 positive pun instances: 194 Homophonic Puns and 251 Homographic Puns. For each positive instance, we generate two types of adversarial negatives, yielding a total of 890 negative samples. 

\begin{table}[h]
\centering
\small
\resizebox{\linewidth}{!}{%
\begin{tabular}{lccc} % 修改列定义，增加一列
\toprule
\textbf{Category} & \textbf{Homophonic} & \textbf{Homographic} & \textbf{Total} \\ % 增加表头
\midrule
Positive Samples & 194 & 251 & 445 \\
\midrule
\multicolumn{4}{l}{\textit{Negative Samples:}} \\ % 修改跨列数量为4
\quad Explicative Substitution (ES) & 194 & 251 & \textbf{445} \\
\quad Random Substitution (RS) & 194 & 251 & \textbf{445} \\
\midrule
Total Negatives & 388 & 502 & \textbf{890} \\
\midrule
\textbf{Total (Pos + Neg)} & \textbf{582} & \textbf{753} & \textbf{1335} \\
\bottomrule
\end{tabular}
}
% \vspace{0.5em}
\caption{Dataset statistics for \textsc{MultiPun}.}
\label{tab:dataset_stats}
\vspace{-10px}
\end{table}

\section{Linguistic Filtering Criteria}
\label{appendix:filtering}

\subsection{WordNet Lexical File Categories}

Table~\ref{tab:lexname_categories} lists the WordNet lexical file categories used in our filtering pipeline. We retain only nouns from \textit{visual} categories (e.g., \texttt{noun.animal}, \texttt{noun.artifact}) to ensure imageability, while filtering out abstract concepts.

\begin{table}[ht]
\centering
\begin{center}
\setlength{\tabcolsep}{4pt}
% 使用 resizebox 确保表格宽度自适应文档宽度
\resizebox{\linewidth}{!}{%
\begin{tabular}{ccc} % 调整最后一列宽度以容纳详细描述
\toprule
\textbf{Category} & \textbf{Lexname} & \textbf{Description} \\
\midrule

% Visual Categories (Retained)
\multirow{8}{*}{\textbf{Visual}} 
 & \texttt{noun.animal} & Animals and distinct biological organisms \\
 \cmidrule(l){2-3}
 & \texttt{noun.artifact} & Man-made objects, tools, and instruments \\
 \cmidrule(l){2-3}
 & \texttt{noun.body} & Body parts (used restrictively) \\
 \cmidrule(l){2-3}
 & \texttt{noun.food} & Edible substances and dishes \\
 \cmidrule(l){2-3}
 & \texttt{noun.object} & Natural inanimate objects (e.g., stones) \\
 \cmidrule(l){2-3}
 & \texttt{noun.plant} & Vegetation and botanical entities \\
\midrule

% Abstract Categories (Filtered)
\multirow{15}{*}{\textbf{Abstract}} 
 & \texttt{noun.location} & Spatial locations and regions \\
 \cmidrule(l){2-3}
 & \texttt{noun.substance} & Substances and bodies of matter \\
 \cmidrule(l){2-3}
 & \texttt{noun.act} & Actions, events, and processes \\
 \cmidrule(l){2-3}
 & \texttt{noun.attribute} & Qualities, properties, and attributes \\
 \cmidrule(l){2-3}
 & \texttt{noun.cognition} & Cognitive processes and contents \\
 \cmidrule(l){2-3}
 & \texttt{noun.communication} & Communicative processes and contents \\
 \cmidrule(l){2-3}
 & \texttt{noun.feeling} & Emotions, feelings, and sensations \\
 \cmidrule(l){2-3}
 & \texttt{noun.motive} & Goals, motives, and wants \\
 \cmidrule(l){2-3}
 & \texttt{noun.quantity} & Quantities, units, and measurements \\
 \cmidrule(l){2-3}
 & \texttt{noun.time} & Temporal points and periods \\
 \cmidrule(l){2-3}
 & \texttt{noun.Tops} & Top-level unique beginners \\

\bottomrule
\end{tabular}%
}
\end{center}
\vspace{-5px}
\caption{Classification of WordNet Lexnames into Visual Anchor Categories (retained) and Abstract Categories (filtered).} 
\vspace{-5px}
\label{tab:lexname_categories}
\end{table}

\subsection{Frequency Thresholds}
To ensure common usage, we apply specific Zipf frequency thresholds. For homophonic puns, we require a frequency greater than 3.0 for both $w_p$ and $w_a$. For homographic puns, we impose a higher threshold of 3.8 for $w_p$ to ensure recognizability given that both senses share the same word form.

\begin{algorithm}[t]
\caption{Diversity Filtering} \label{alg:sentence_embedding_filtering}
\begin{algorithmic}[1]
\STATE \textbf{Input:} candidate dataset $\mathcal{D}=\{d_i\}_{i=1}^{N}$, target size $k$ ($k < N$), embedding function $\mathrm{EMB}$
\STATE \textbf{Output:} filtered diverse subset $\mathcal{D}' \subseteq \mathcal{D}$ with $|\mathcal{D}'|=k$, minimum pairwise distance $d_{\min}$
\vspace{4pt}
\STATE Compute sentence embeddings $e_i \leftarrow \mathrm{EMB}(d_i)$ for all $i=1,\dots,N$
\STATE Construct pairwise cosine distance matrix $\mathbf{D}\in\mathbb{R}^{N\times N}$ by $D_{ij} \;=\; 1-\frac{e_i^\top e_j}{\lVert e_i\rVert\,\lVert e_j\rVert}, D_{ii}\leftarrow +\infty$
       \hfill $\triangleright$ Lower $D_{ij}$ indicates higher semantic similarity
       \vspace{-12px}
\STATE Initialize active candidate set $\mathcal{S}\leftarrow\{1,\dots,N\}$
\vspace{-12px}
\FOR{iteration $t=1$ \textbf{to} $N-k$}
    \STATE Identify the most similar pair $(i,j)\leftarrow \mathop{\mathrm{arg\,min}}_{p\neq q,\;p,q\in\mathcal{S}} D_{pq}$
           \hfill $\triangleright$ Find the closest pair with minimum distance
    \STATE Calculate redundancy scores for the closest pair $\phi_i=\sum_{v\in\mathcal{S}} D_{iv},
           \phi_j=\sum_{v\in\mathcal{S}} D_{jv}$
           \hfill $\triangleright$ Lower $\phi$ indicates higher centrality
    \STATE Select the more redundant candidate:$u \leftarrow \mathop{\mathrm{arg\,min}}\{\phi_i,\phi_j\}$
           \hfill $\triangleright$ Choose the candidate closer to the remaining set
    \STATE Update active set: $\mathcal{S} \leftarrow \mathcal{S} \setminus \{u\}$
           \hfill $\triangleright$ Remove the more redundant candidate
\ENDFOR
\STATE Construct final subset $\mathcal{D}' \leftarrow \{d_i \mid i\in\mathcal{S}\}$
\STATE Compute diversity $d_{\min}\leftarrow \min_{i\neq j,\; i,j\in\mathcal{S}} D_{ij}$
\RETURN $\mathcal{D}',\, d_{\min}$
\end{algorithmic}
\end{algorithm}

\subsection{Diversity Filtering}
\label{app:diversity_filtering}

We use the deterministic filtering process outlined in Algorithm~\ref{alg:sentence_embedding_filtering} to select the final $k$ items. Given the candidate dataset $\mathcal{D}$ of $N$ items, we first compute the sentence embeddings $e_i = \text{EMB}(d_i)$ for all items using \texttt{text-embedding-3-large}, where $d_i$ is the ground-truth rationale text. We then construct the pairwise cosine distance matrix $\mathbf{D}$. The algorithm iteratively prunes the dataset $N-k$ times. In each iteration, it identifies the most similar pair of candidates $(i,j)$ in the active set $\mathcal{S}$ (Line 6). To decide which candidate to remove, it calculates a redundancy score $\phi$ for both $i$ and $j$, defined as the sum of distances to all other active candidates (Line 8). The candidate with the smaller $\phi$ is deemed more central or more redundant and is removed from $\mathcal{S}$ (Lines 10 and 12). By iteratively removing the most redundant candidate from each closest pair, this process ensures that semantic outliers are preserved~\cite{ENCODER,ReTrack,HABIT,lan2025mappo}, and the final set of $k$ items maintains maximum conceptual diversity and coverage~\cite{wu2026tsembed,wu2024heterogeneity,wu2025breaking}.

\section{Generation Prompts}
\label{appendix:prompts}

This section provides the prompt templates used for generating positive pun samples and adversarial negative samples in the \textsc{MultiPun} dataset.

\subsection{Positive Sample Generation}

\subsubsection{Homophonic Pun Creation Prompt}

\begin{PromptBox}{Creative Prompt for Homophonic Puns}
\# Role

You are an expert in multimodal humor. Your task is to generate visual pun data based on \textbf{Homophones} (words that sound the same but have different meanings and spellings).

\# Task Definition

I will provide you with two words:

1.  \textbf{Word A (Visual Object):} The word that determines the visual appearance ($S_p$).

2.  \textbf{Word B (Hidden Context):} The word that determines the behavior/action ($S_a$).

You need to generate:

1.  \textbf{Image Description:} Description of Object A acting out the meaning of Word B.

2.  \textbf{Caption:} A sentence containing Word A, but implying Word B.

3.  \textbf{Interpretation:} An analysis of the pun.

\# Example

\textbf{Input:}

* \textbf{Word A:} pear: sweet juicy gritty-textured fruit available in many varieties

* \textbf{Word B:} pair: two items of the same kind

\textbf{Output:}

\textbf{Image Description:} Two cartoon pears holding hands and smiling happily at each other.

\textbf{Caption:} We make a great pear.

\textbf{Interpretation:} Visual depicts two pears (literal object, $S_p$) holding hands like a romantic pair (figurative behavior, $S_a$). The caption exploits the homophonic relationship between 'pear' ($w_p$) and 'pair' ($w_a$), creating humor through sound similarity between different meanings.

\# Current Input

* \textbf{Word A:} \textcolor{blue}{\textbf{[Insert Word A, e.g., Chili: a small hot-tasting pod of a variety of capsicum]}}

* \textbf{Word B:} \textcolor{blue}{\textbf{[Insert Word B, e.g., Chilly: uncomfortably cool or cold]}}

\# Output
\end{PromptBox}

\subsubsection{Homographic Pun Creation Prompt}

\begin{PromptBox}{Creative Prompt for Homographic Puns}
\# Role

You are an expert in multimodal humor. Your task is to generate visual pun data based on \textbf{Homographic Puns} (a single word with multiple meanings in the same spelling).

\# Task Definition

I will provide you with one word and its two distinct definitions:

1.  \textbf{The Word:} The lexical item used in the caption.

2.  \textbf{Definition 1 (Visual Object):} The literal/concrete meaning that determines the physical appearance of the object ($S_p$).

3.  \textbf{Definition 2 (Hidden Context):} The figurative behavior/state meaning that determines the behavior, action, or setting ($S_a$).

You need to generate:

1.  \textbf{Image Description:} A description of the object from Definition 1 performing the action or situated in the context of Definition 2.

2.  \textbf{Caption:} A witty sentence using "The Word", where the sentence structure strongly implies Definition 2.

3.  \textbf{Interpretation:} A concise explanation of the pun mechanism.

\# Example

\textbf{Input:}

* \textbf{The Word:} fan

* \textbf{Definition 1:} a device for creating a current of air by movement of a surface or surfaces

* \textbf{Definition 2:} an ardent follower and admirer

\textbf{Output:}

* \textbf{Image Description:} A large electric floor fan in a stadium seat, holding a foam finger and cheering loudly.

* \textbf{Caption:} I'm your biggest fan.

* \textbf{Interpretation:} Visual shows a cooling fan (literal object, $S_p$); caption uses 'fan' as admirer (figurative behavior, $S_a$), creating a homographic pun where the same word embodies both meanings.

\# Current Input

* \textbf{The Word:} \textcolor{blue}{\textbf{[Insert Word Here]}}

* \textbf{Definition 1 (Visual Object):} \textcolor{blue}{\textbf{[Insert Literal Definition Here]}}

* \textbf{Definition 2 (Hidden Context):} \textcolor{blue}{\textbf{[Insert Abstract/Contextual Definition Here]}}

\# Output
\end{PromptBox}

\subsection{Adversarial Negative Sample Generation}

\subsubsection{Explicative Substitution}

\begin{PromptBox}{Explicative Substitution Generation}
You are a data augmentation expert. Given the following pun, generate an Explicative Substitution variant:

Original Caption: \textcolor{blue}{\textbf{\{caption\}}}

Pun Word ($w_p$): \textcolor{blue}{\textbf{\{word\}}}

Hidden Meaning ($S_a$): \textcolor{blue}{\textbf{\{meaning\}}}

Task: Replace $w_p$ with an EXPLICIT STATEMENT of the hidden meaning $S_a$.

Constraints:

- Do NOT use $w_p$ or $w_a$ directly

- Use paraphrases or synonyms to express $S_a$

- Adjust grammar if needed for naturalness

- Prefer single-word replacements when possible

Example:

Original: ``We make a great pear.''

Hidden Meaning: romantic couple

Output: ``We make a great romantic couple.''
\end{PromptBox}

\subsubsection{Random Substitution}

\begin{PromptBox}{Random Substitution Generation}
You are a data augmentation expert. Given the following pun, generate a Random Substitution variant:

Original Image Prompt: \textcolor{blue}{\textbf{\{visual description\}}}

Original Caption: \textcolor{blue}{\textbf{\{caption\}}}

Pun Word ($w_p$): \textcolor{blue}{\textbf{\{word\}}}

Task:

1. Select a RANDOM concrete noun (e.g., chair, banana, bicycle, umbrella, book) that is SEMANTICALLY UNRELATED to the original pun context

2. Replace the main object in the image prompt with this random entity

3. Replace $w_p$ in the caption with the same random entity

4. Keep the same action/context structure

Constraints:

- The random entity must be a concrete, visualizable noun

- Must be completely unrelated to original pun

- Do NOT reuse common examples (vary your selection)

Example:

Original Visual: ``Two cartoon pears holding hands...''

Original Caption: ``We make a great pear.''

Random Entity: banana

New Visual: ``Two cartoon bananas holding hands...''

New Caption: ``We make a great banana.''
\end{PromptBox}

\section{Human Verification Protocol}
\label{appendix:human_verification}
We recruited three graduate students from our institution with prior experience in NLP research~\cite{liu2026humanstudy}. All participants were aged 20-28 years and consisted of two male and one female doctoral students in computer science. Participants were compensated at \$25 USD/hour (approximately 8 hours per participant) and provided informed consent. All annotations were anonymized and used only for academic research.
All generated samples (positive and negative) undergo human verification. Three annotators independently evaluate each sample based on:

\begin{enumerate}[nosep,leftmargin=11pt]
    \item \textbf{Image Quality:} Is the visual content clear, non-distorted, and depicts the intended object?
    \item \textbf{Visual-Textual Coherence:} For positive samples, does the visual content coherently connect to the text description? For negative samples, is the intended disruption (ES/RS) clearly present?
    \item \textbf{Ambiguity Presence:} For positive samples, is there genuine dual-layer semantics? For negative samples, is the ambiguity properly resolved?
    \item \textbf{Naturalness:} Are the caption and visual scenario natural and plausible?
\end{enumerate}

Samples are retained if at least 2 out of 3 annotators agree on acceptance. Rejected samples are either regenerated with refined prompts or discarded. The inter-annotator agreement (Fleiss' Kappa) across all samples is 0.78, indicating substantial agreement.

\section{Evaluation Suite Task Descriptions}
\label{appendix:eval_prompts}

Our evaluation suite comprises three recognition tasks with progressive levels of structural guidance: \textbf{Detection}, \textbf{Localization}, and \textbf{Explanation}. For each task, we use two prompt variants to separate true reasoning from affirmative language bias~\cite{liu2025cobra}: (1) \textit{biased-to-pun} prompt that asks whether the given context is a pun, and (2) \textit{biased-to-non-pun} prompt that asks whether the given context is not a pun. The key difference is in the task description and output order, while the definitions and requirements remain identical.

All experiments are run three times, and the reported results are averages. All baselines follow their official implementations.

\subsection{Detection}

This task asks for binary judgment (pun or not). We provide two variants: one without formal definitions and one with formal definitions and notation.

\subsubsection{Pun Detection}

\begin{PromptBox}{Detection without Definitions (Biased-to-Pun)}
You are an expert linguist specializing in Multimodal Puns.

\textbf{Task Description}

Analyze the provided image and caption to determine if they constitute a \textbf{Multimodal Pun}.

\textbf{Input Data}

Caption: \textcolor{blue}{\textbf{\{caption\}}}

\textbf{Output Requirements}

Output ONLY a JSON object:

\{"is\_pun": true/false\}

IMPORTANT: Output ONLY the JSON object, no additional text or explanation.
\end{PromptBox}

\textit{Note: The biased-to-non-pun variant changes the task description to "determine if they constitute a \textbf{Non-Pun} (not a pun)" and adds "Note: Answer true if it is a pun, false if it is a non-pun."}

\subsection{Pun Localization}

This task requires first judging and explicitly identifying words $w_p$ and $w_a$.

\begin{PromptBox}{Localization (Biased-to-Pun)}
You are an expert linguist specializing in Multimodal Puns.

\textbf{Task Description}

Analyze the provided image and caption to determine if they constitute a \textbf{Multimodal Pun}. If yes, categorize the pun type and extract ONLY the word pair ($w_p$ and $w_a$).

\textbf{Definitions}

\begin{enumerate}[nosep]
\item \textbf{Homophonic Pun:} The caption contains a word that sounds like another word with \textbf{different spelling and meaning}.
\begin{itemize}[nosep]
\item $w_p$: The word \textbf{actually appearing in the caption}
\item $w_a$: The hidden word it sounds like (different spelling/meaning)
\item Example: ``pear'' (in caption) sounds like ``pair'' (hidden meaning)
\end{itemize}

\item \textbf{Homographic Pun:} The caption contains a word with \textbf{two distinct meanings in the same spelling}.
\begin{itemize}[nosep]
\item $w_p$ and $w_a$ are the \textbf{same word} appearing in the caption (both should be identical)
\item Example: ``fan'' means both ``cooling device'' and ``enthusiast''
\end{itemize}
\end{enumerate}

\textbf{Input Data}

Caption: \textcolor{blue}{\textbf{\{caption\}}}

\textbf{Output Requirements}

If it is NOT a pun:

\{"is\_pun": false\}

If it IS a pun:

\{
  "is\_pun": true,
  "type": "<Homophonic or Homographic>",
  "tuple": \{
    "wp": "<The EXACT word appearing in the caption>",
    "wa": "<The hidden/alternative word>"
  \}
\}

IMPORTANT: Output ONLY the JSON object with the fields shown above. Do NOT include semantic definitions ($S_p$ or $S_a$). Only provide the word pair (wp and wa). No additional text or explanation.
\end{PromptBox}

\subsection{Pun Explanation}

This task requires judging, providing a rationale that explains why it's a pun, and extracting the full tuple $\langle w_p, w_a, S_p, S_a \rangle$.

\begin{PromptBox}{Explanation (Biased-to-Pun)}
You are an expert linguist specializing in Multimodal Puns.

\textbf{Task Description}

Analyze the provided image and caption to determine if they constitute a \textbf{Multimodal Pun}. If yes, categorize the pun type and extract the linguistic components following the formal notation $P = \langle w_p, w_a, S_p, S_a \rangle$.

\textbf{CRITICAL RULE: What is a Multimodal Pun?}

A multimodal pun MUST satisfy ALL of the following conditions:
\begin{enumerate}[nosep]
\item \textbf{The pun word MUST explicitly appear in the caption text}
\item \textbf{This word must create dual meanings through either:}
\begin{itemize}[nosep]
\item \textbf{Phonetic similarity} (sounds like another word with different spelling/meaning)
\item \textbf{Lexical polysemy} (same spelling but two distinct meanings)
\end{itemize}
\item \textbf{Visual-linguistic coupling:} The image fuses a literal object ($S_p$) with a figurative behavior/state ($S_a$), while the text unifies them through the pun word
\end{enumerate}

\textbf{IMPORTANT:} If the caption does not contain the pun word, or if the visual and textual meanings are not genuinely linked, it is NOT a multimodal pun.

\textbf{Definitions}

\begin{enumerate}[nosep]
\item \textbf{Homophonic Pun:} Exploits sound similarity between words with \textbf{different spelling and meaning}.
\begin{itemize}[nosep]
\item $w_p$: The word \textbf{actually appearing in the caption}
\item $w_a$: The hidden word it sounds like (different spelling/meaning)
\item $S_p$: The literal/concrete object depicted in the image
\item $S_a$: The figurative behavior/state associated with the alternative word
\item Example: ``We make a great pear'' --- image shows pears ($S_p$) holding hands like a romantic pair ($S_a$)
\end{itemize}

\item \textbf{Homographic Pun:} Exploits dual meanings of a word with \textbf{the same spelling}.
\begin{itemize}[nosep]
\item $w_p$ and $w_a$ are the \textbf{same word} appearing in the caption
\item $S_p$: The concrete/literal sense depicted visually in the image
\item $S_a$: The figurative/abstract sense implied by the textual context
\item Example: ``I'm a big fan of yours'' --- image shows a cooling fan ($S_p$) cheering like an enthusiast ($S_a$)
\end{itemize}
\end{enumerate}

\textbf{Input Data}

Caption: \textcolor{blue}{\textbf{\{caption\}}}

\textbf{Analysis Steps}

\begin{enumerate}[nosep]
\item \textbf{First, identify if there is a word in the caption that could have dual meanings}
\item \textbf{Check if one meaning relates to the image and another to the text context}
\item \textbf{Only if BOTH conditions are met, classify as a pun}
\end{enumerate}

\textbf{Output Requirements}

\textbf{Condition A: If it is NOT a pun:}

Output exactly this JSON:

\{"is\_pun": false\}

\textbf{Condition B: If it IS a pun:}

The pun word MUST be present in the caption. Output:

\{
  "is\_pun": true,
  "type": "<Homophonic or Homographic>",
  "explanation": "<Brief explanation of how the pun creates humor through visual-linguistic interplay>",
  "tuple": \{
    "wp": "<The EXACT word appearing in the caption that creates the pun>",
    "wa": "<The alternative word: different spelling if Homophonic, same spelling if Homographic>",
    "Sp": "<The literal/concrete meaning shown in the image>",
    "Sa": "<The figurative/abstract meaning implied by context>"
  \}
\}

IMPORTANT: Output ONLY the JSON object, no additional text or explanation.
\end{PromptBox}

\section{Pun-CoT: Enhanced Prompt with Three-Stage Verification}
\label{appendix:pun_cot}

To address the hallucination errors identified in our error analysis (Section~\ref{sec:recognize}), we propose \textbf{Pun-CoT} (Pun-aware Chain-of-Thought), an enhanced prompt that enforces a structured three-stage verification process. This method is designed to mitigate four common error patterns: pun keyword hallucination, phonetic hallucination, semantic hallucination, and visual object hallucination.

\begin{PromptBox}{Pun-CoT Enhanced Prompt (Biased-to-Pun)}
You are an expert linguist specializing in Multimodal Puns.

\textbf{Task Description}

Analyze the provided image and caption to determine if they constitute a \textbf{Multimodal Pun}. Use a structured three-stage verification process to avoid common errors.

\textbf{Formal Definition}

A multimodal pun is represented as $P = \langle w_p, w_a, S_p, S_a \rangle$ where:
\begin{itemize}[nosep]
\item $w_p$: The pun word \textbf{explicitly appearing in the caption}
\item $w_a$: The alternative word (hidden meaning)
\item $S_p$: The literal/concrete object sense (depicted visually in the image)
\item $S_a$: The figurative behavior/state sense (implied by textual context)
\end{itemize}

\textbf{Pun Types}

\begin{enumerate}[nosep]
\item \textbf{Homophonic Pun:} Exploits sound similarity between words with \textbf{different spelling and meaning}
\begin{itemize}[nosep]
\item Example: ``pear'' (in caption) sounds like ``pair'' (hidden meaning)
\item Image shows pears (literal object) holding hands like a romantic pair (figurative behavior)
\end{itemize}

\item \textbf{Homographic Pun:} Exploits dual meanings of a word with \textbf{the same spelling}
\begin{itemize}[nosep]
\item Example: ``fan'' means both ``cooling device'' and ``enthusiast''
\item Image shows a fan device (literal object) cheering like an enthusiast (figurative behavior)
\end{itemize}
\end{enumerate}

\textbf{CRITICAL THREE-STAGE VERIFICATION}

\textbf{STAGE 1: Visual Grounding (Prevent Visual Object Hallucination)}
\begin{itemize}[nosep]
\item First, describe EXACTLY what visual object you see in the image
\item DO NOT infer objects based on text context
\item DO NOT assume objects that are not visually present
\item Example: If you see apples, do NOT call them ``dates'' even if the text mentions ``date''
\end{itemize}

\textbf{STAGE 2: Lexical Anchoring (Prevent Pun Keyword Hallucination)}
\begin{itemize}[nosep]
\item Identify the EXACT words in the caption text
\item DO NOT mentally replace words with idiom components
\item Example: If caption says ``I'm your biggest lamp'', do NOT treat it as if it says ``fan''
\item List all potential pun candidates from the ACTUAL caption words
\end{itemize}

\textbf{STAGE 3: Cross-Modal Verification (Prevent Phonetic/Semantic Hallucination)}

For each potential pun word, verify:

a) \textbf{Phonetic Bridge (for Homophonic):} Do $w_p$ and $w_a$ ACTUALLY sound similar?
\begin{itemize}[nosep]
\item REJECT if phonetically distinct (e.g., ``banana'' does NOT sound like ``soul'')
\item Require genuine phonetic similarity
\end{itemize}

b) \textbf{Semantic Bridge (for Homographic):} Does the word have TWO established meanings?
\begin{itemize}[nosep]
\item REJECT if forcing meanings onto unrelated words
\item Example: ``banana'' does NOT have a meaning related to ``pair'' or ``couple''
\end{itemize}

c) \textbf{Visual-Textual Link:} Does the visual object connect to text via valid pun mechanism?
\begin{itemize}[nosep]
\item For Homophonic: Visual shows $S_p$ (literal object of $w_p$), text implies $S_a$ (figurative behavior of $w_a$)
\item For Homographic: Same word connects both the literal visual sense and figurative textual sense
\item REJECT weak or fabricated connections
\end{itemize}

\textbf{Input Data}

Caption: \textcolor{blue}{\textbf{\{caption\}}}

% \textbf{Verification Checklist (Execute in Order)}

% \begin{enumerate}[nosep]
% \item $\checkmark$ What object do I see in the image? (Visual Grounding)
% \item $\checkmark$ What are the exact words in the caption? (Lexical Anchoring)
% \item $\checkmark$ Is there a valid phonetic or semantic bridge? (Cross-Modal Verification)
% \item $\checkmark$ Does the pun word appear in the caption? (Final Check)
% \end{enumerate}

\textbf{Output Requirements}

If it is NOT a pun (failed any verification stage):

\{"is\_pun": false\}

If it IS a pun (passed all verification stages):

\{
  "is\_pun": true,
  "type": "<Homophonic or Homographic>",
  "explanation": "<Brief explanation of the verified pun mechanism>",
  "tuple": \{
    "wp": "<The EXACT word appearing in the caption>",
    "wa": "<The alternative word: different spelling if Homophonic, same spelling if Homographic>",
    "Sp": "<The literal/concrete meaning shown in the image>",
    "Sa": "<The figurative/abstract meaning implied by context>"
  \}
\}

IMPORTANT:
\begin{itemize}[nosep]
\item Execute ALL three verification stages before making judgment
\item Be conservative: when in doubt, classify as NOT a pun
\item The pun word MUST explicitly appear in the caption
\item Output ONLY the JSON object, no additional text
\end{itemize}
\end{PromptBox}

\section{Model Configuration}
\label{app:models}

We evaluate a total of 11 VLMs. Tables~\ref{tab:closed_source_models} and~\ref{tab:open_source_models} provide comprehensive overviews of all evaluated models and their configurations.

\subsection{Closed-Source VLMs}

Table~\ref{tab:closed_source_models} presents the configuration details for closed-source models accessed via API.

\begin{table}[h]
\centering
\small
% \resizebox{0.8\linewidth}{!}{
\begin{tabular}{ll}
\toprule
\textbf{Model} & \textbf{API Version} \\
\midrule
\multicolumn{2}{c}{\textit{OpenAI Family}} \\
\midrule
GPT-5.1 & gpt-5.1 \\
GPT-4o & gpt-4o-2024-08-06 \\
\midrule
\multicolumn{2}{c}{\textit{Google Gemini Family}} \\
\midrule
Gemini-3-Pro & gemini-3-pro-preview \\
\midrule
\multicolumn{2}{c}{\textit{Anthropic Family}} \\
\midrule
Claude-Sonnet-4.5 & claude-sonnet-4-5-20250929 \\
\bottomrule
\end{tabular}
% }
\caption{Closed-source VLM configurations.}
\vspace{-10px}
\label{tab:closed_source_models}
\end{table}

\subsection{Open-Source VLMs}

Table~\ref{tab:open_source_models} presents the configuration details for open-source models. All models are evaluated using their officially released checkpoints from Hugging Face by hosting the model on a vLLM server. 

\begin{table}[h]
\centering
\small
\setlength\tabcolsep{2pt}
\resizebox{\linewidth}{!}{
\begin{tabular}{llc}
\toprule
\textbf{Model} & \textbf{Checkpoint} & \textbf{Type} \\
\midrule
\multicolumn{3}{c}{\textit{Meta Llama-4 Family}} \\
\midrule
Llama-4-Scout-17B & meta-llama/Llama-4-Scout-17B-16E-Instruct & Instruct \\
\midrule
\multicolumn{3}{c}{\textit{Alibaba Qwen3-VL Family}} \\
\midrule
Qwen3-VL-8B-Instruct & Qwen/Qwen3-VL-8B-Instruct & Instruct \\
Qwen3-VL-30B-A3B-Instruct & Qwen/Qwen3-VL-30B-A3B-Instruct & Instruct \\
Qwen3-VL-8B-Thinking & Qwen/Qwen3-VL-8B-Thinking & Reasoning \\
Qwen3-VL-30B-A3B-Thinking & Qwen/Qwen3-VL-30B-A3B-Thinking & Reasoning \\
\midrule
\multicolumn{3}{c}{\textit{LLaVA Family}} \\
\midrule
LLaVA-V1.6-Vicuna-13B & liuhaotian/llava-v1.6-vicuna-13b & Instruct \\
\bottomrule
\end{tabular}
}
\caption{Open-source VLM configurations.}
\vspace{-10px}

\label{tab:open_source_models}
\end{table}

\subsection{Hardware}

All open-source models are evaluated on two NVIDIA A100 80GB GPUs. Closed-source models are accessed via their official APIs.

\section{Additional Results}
\label{app:additional_results}

\begin{figure}[t]
\begin{center}
\centerline{\includegraphics[width=\linewidth]{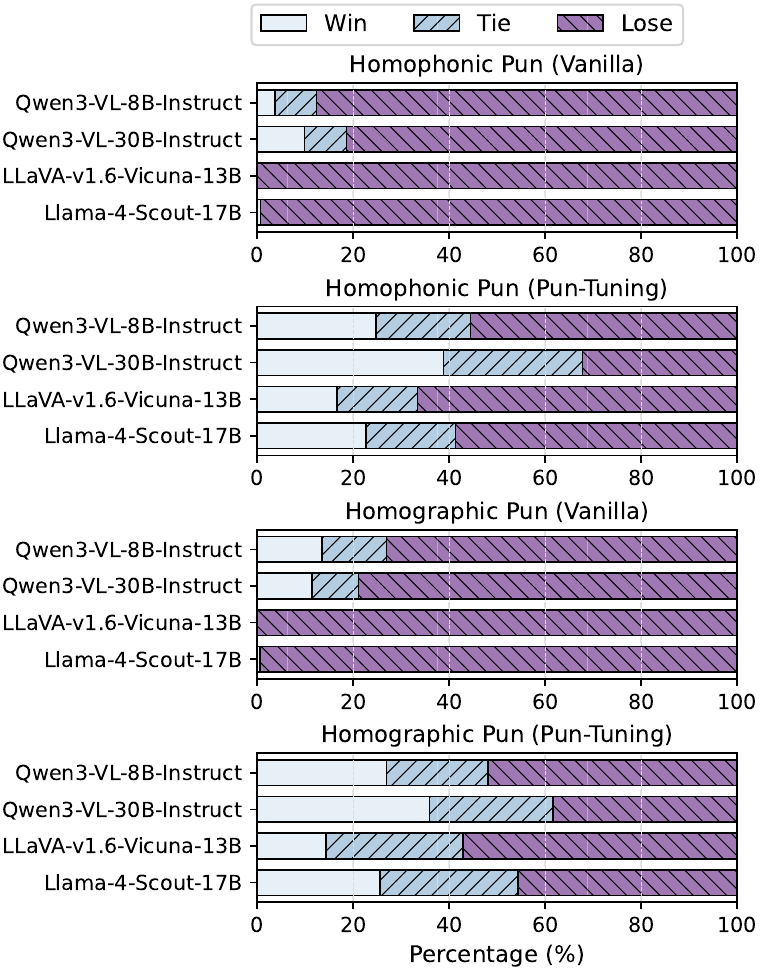}}
\vspace{-5px}
\caption{Pairwise comparison for pun explanations before and after Pun-Tuning.} 
\label{fig:pairwise_tuning}
\end{center}
\vspace{-25px}
\end{figure}

Figure~\ref{fig:pairwise_tuning} shows the pairwise comparison for pun explanations before and after Pun-Tuning.

\section{Pun-Tuning Implementation Details}
\label{app:pun_tuning_details}

\subsection{Dataset Splits}

We split the dataset ensuring no test samples leak into training. The 194 homophonic puns are divided into 97 training and 97 test samples; the 251 homographic puns are split into 125 training and 126 test samples. Negative samples maintain a 2:1 ratio with positive samples (each positive sample paired with 2 negatives: one Explicative Substitution and one Random Substitution). Table~\ref{tab:pun_tuning_splits} shows the complete breakdown.

\begin{table}[t]
\centering
\small
\resizebox{0.8\linewidth}{!}{%
\begin{tabular}{lcccc}
\toprule
\textbf{Category} & \textbf{Pun Type} & \textbf{Train} & \textbf{Test} & \textbf{Total} \\
\midrule
\multirow{2}{*}{\textbf{Positive}} 
 & Homophonic & 97 & 97 & 194 \\
 \cmidrule(l){2-5}
 & Homographic & 125 & 126 & 251 \\
\midrule
\multirow{2}{*}{\textbf{Negative}} 
 & Homophonic & 194 & 194 & 388 \\
 \cmidrule(l){2-5}
 & Homographic & 250 & 252 & 502 \\
\midrule
\textbf{Total} & & \textbf{666} & \textbf{669} & \textbf{1335} \\
\bottomrule
\end{tabular}%
}
\caption{Dataset splits for Pun-Tuning experiments.}
\label{tab:pun_tuning_splits}
\end{table}

\subsection{Hyperparameters}

We fine-tune three open-source models (Qwen3-VL-8B-Instruct, Qwen3-VL-30B-A3B-Instruct, and LLaVA-V1.6-Vicuna-13B) with batch size 4 per A100 GPU, learning rate 2e-5, AdamW optimizer, linear warmup (100 steps) followed by cosine decay, weight decay 0.01, gradient clipping (max norm 1.0), and FP16 mixed precision for 3 epochs. Training uses both \textit{biased-to-pun} and \textit{biased-to-non-pun} prompt variants. Evaluation is performed on the held-out test set (669 samples) across all three tasks~\cite{wu2025memory,wudevelopmental,wu2025elastic}.

\section{Software Packages}
\label{app:packages_parameters}

We use the following Python packages: NLTK (version 3.9.2) for WordNet access and lemmatization, and the pronouncing package (version 0.2.0) for CMU Pronouncing Dictionary access.

\section{The Use of Large Language Models} 
We utilize LLMs to assist with language and code polishing, as well as error checking, during the preparation of this manuscript. The content, ideas, and scientific contributions remain entirely our own, and all substantive intellectual work is conducted by the authors.

% \clearpage
% \onecolumn

\section{Failure Cases}
\subsection{Error Analysis of Negative Samples}
\label{app:appendix_negative_cases}

VLMs often fail to correctly reject non-pun distractors, exhibiting systematic hallucination patterns across lexical, phonological, semantic, and visual levels. Table~\ref{tab:appendix_negative_cases} provides qualitative examples of these four error types, demonstrating how models incorrectly interpret negative samples as genuine puns despite the absence of valid pun mechanisms.

\begin{table*}[h]
\centering
\footnotesize 
\definecolor{deepred}{RGB}{200, 0, 0}

% 【修改点】：不再使用 tabularx
% 使用 standard tabular
% p{0.49\textwidth} 表示该列占页面宽度的 49%
% 两个 0.49 加起来是 0.98，剩下的空间留给中间的缝隙
\resizebox{\linewidth}{!}{
\begin{tabular}{p{0.49\textwidth} p{0.49\textwidth}}
\toprule

\multicolumn{2}{l}{\color{blue} \bf Pun Keyword Hallucination}   \\
\midrule
\bf Homographic Pun & \bf  Random Substitution (RS)\\

% ==================== 第 1 行：Pun Keyword Hallucination ====================
% --- 左列 ---
\begin{minipage}[t]{0.28\linewidth} 
    \vspace{0pt} 
    \centering
    \includegraphics[width=\linewidth]{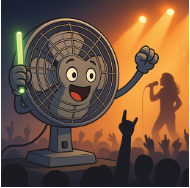} \\
    \vspace{2pt} 
    I'm your biggest \textcolor{deepred}{\textbf{fan}}.
\end{minipage}%
\hfill 
\begin{minipage}[t]{0.68\linewidth} 
    \vspace{0pt} 
    \raggedright \scriptsize 
    \textbf{claude-sonnet-4-5} \newline
    $\bm{w_p}$: fan \newline
    $\bm{S_p}$: A mechanical device with rotating blades that circulates air \newline
    $\bm{w_a}$: fan \newline
    $\bm{S_a}$: An enthusiastic admirer or supporter \newline
    \textbf{Explanation}: The word 'fan' has two meanings: (1) a mechanical device that circulates air, which is visually depicted in the image as an anthropomorphized electric fan, and (2) an enthusiastic supporter or admirer, as indicated by the foam finger and the phrase 'biggest fan' which is an idiom meaning someone's greatest supporter.
\end{minipage}

& % --- 右列 ---
\begin{minipage}[t]{0.28\linewidth}
    \vspace{0pt}
    \centering
    \includegraphics[width=\linewidth]{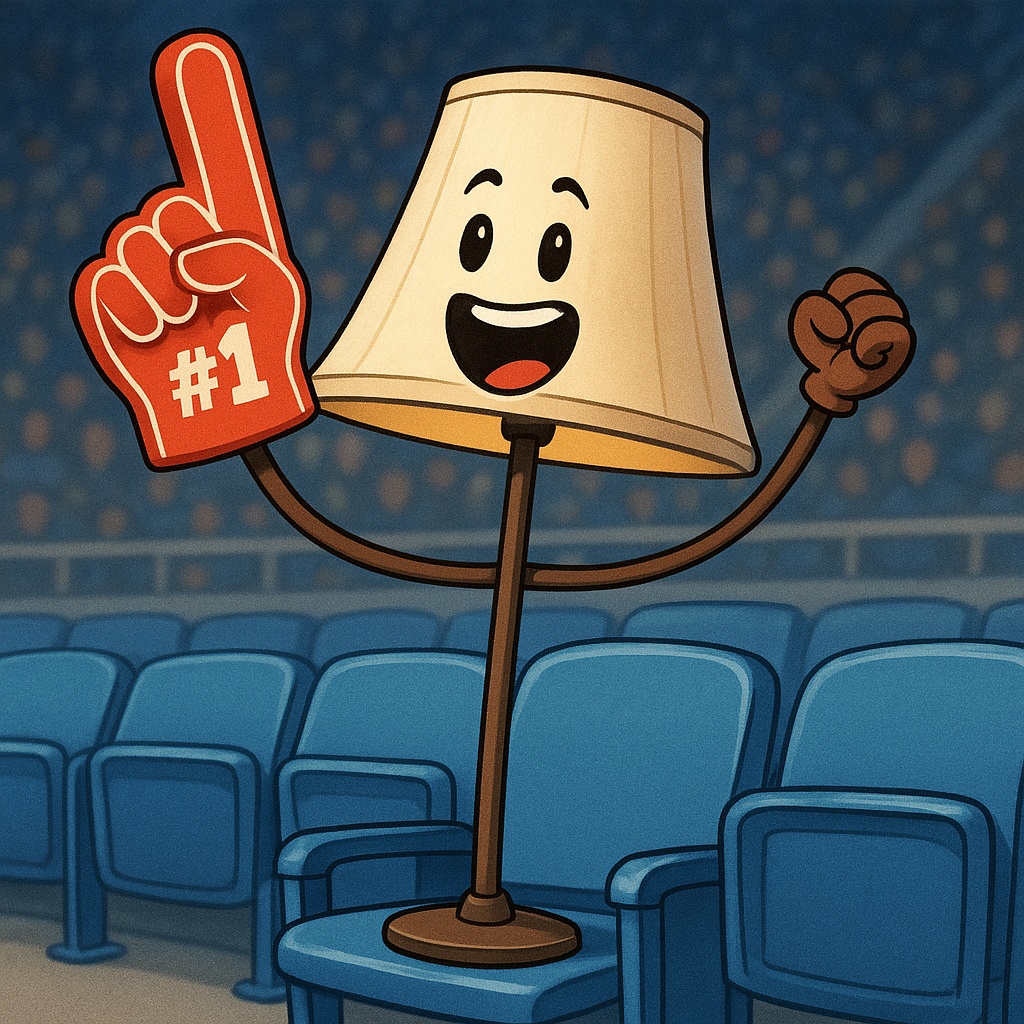} \\
    \vspace{2pt}
    I'm your biggest \textcolor{deepred}{\textbf{lamp}} in the room. 
\end{minipage}%
\hfill
\begin{minipage}[t]{0.68\linewidth}
    \vspace{0pt}
    \raggedright \scriptsize
    \textbf{claude-sonnet-4-5} \newline
    $\bm{w_p}$: lamp \newline
    $\bm{S_p}$: A lighting fixture consisting of a shade and base \newline
    $\bm{w_a}$: fan \newline
    $\bm{S_a}$: An enthusiastic supporter or admirer \newline
    \textbf{Explanation}: The word 'lamp' creates a pun by having two meanings: literally referring to the lamp character in the image, and playing on the phrase 'biggest fan' where 'lamp' sounds like it's substituting for 'fan' in the context of being a supporter at what appears to be a stadium/arena setting.
\end{minipage}
\\ % 行结束

% ==================== 第 2 行：Phonetic Hallucination ====================
\midrule
\multicolumn{2}{l}{\color{blue} \bf Phonetic Hallucination}   \\
\midrule
\bf Homophonic Pun & \bf  Random Substitution (RS)\\

% --- 左列 ---
\begin{minipage}[t]{0.28\linewidth} 
    \vspace{0pt} 
    \centering
    \includegraphics[width=\linewidth]{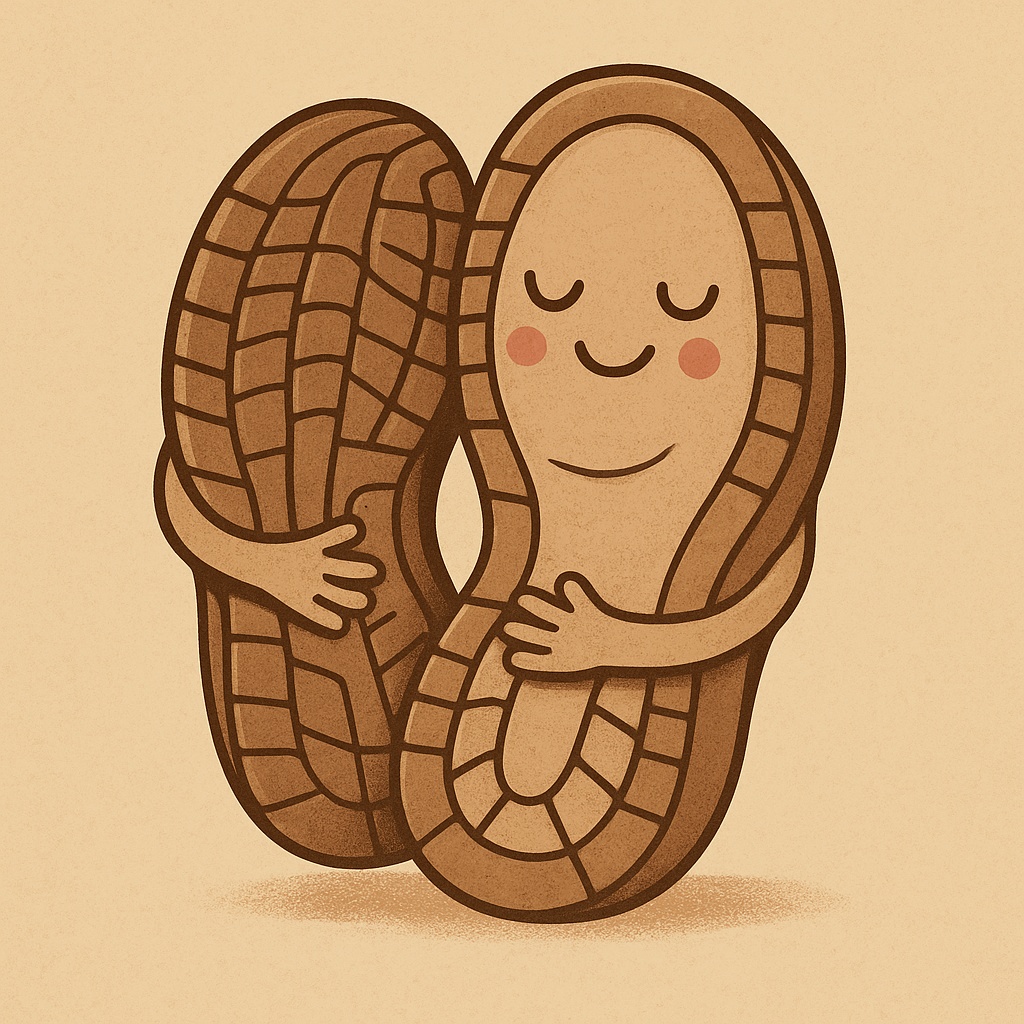} \\
    \vspace{2pt} 
    You are my \textcolor{deepred}{\textbf{sole}} mate.
\end{minipage}%
\hfill 
\begin{minipage}[t]{0.68\linewidth} 
    \vspace{0pt} 
    \raggedright \scriptsize 
    \textbf{Gemini-3-Pro} \newline
    $\bm{w_p}$: sole \newline
    $\bm{S_p}$: The bottom part of a shoe or foot, which is visually depicted as two shoe bottoms hugging. \newline
    $\bm{w_a}$: soul \newline
    $\bm{S_a}$: The spiritual or immaterial part of a human being, used in the phrase 'soul mate' to mean a perfect romantic partner. \newline
    \textbf{Explanation}: The humor relies on the phonetic similarity between 'sole' (the bottom of a shoe, depicted in the image) and 'soul' (as in 'soul mate', a romantic partner). The image shows two shoe soles hugging, literalizing the pun.    
\end{minipage}

& % --- 右列 ---
\begin{minipage}[t]{0.28\linewidth}
    \vspace{0pt}
    \centering
    \includegraphics[width=\linewidth]{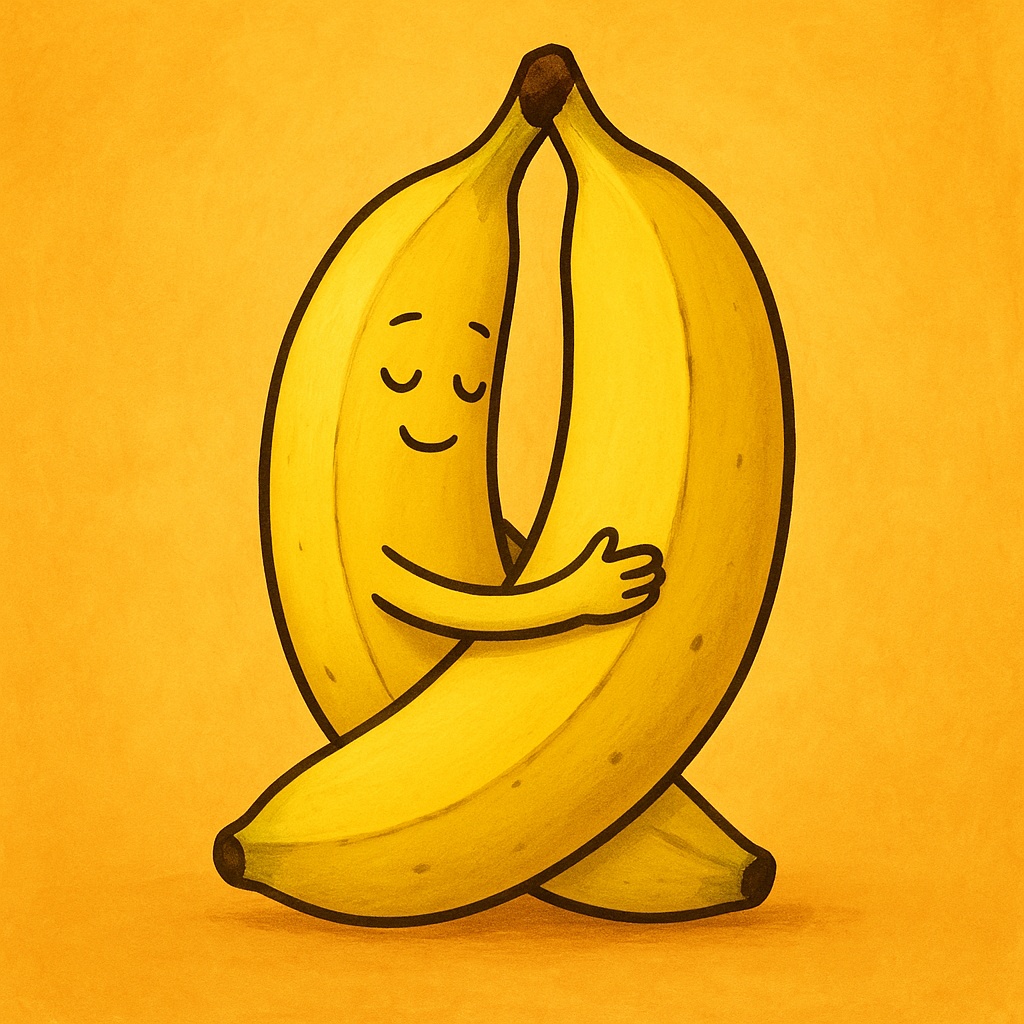} \\
    \vspace{2pt}
    You are my \textcolor{deepred}{\textbf{banana}} mate.
\end{minipage}%
\hfill
\begin{minipage}[t]{0.68\linewidth}
    \vspace{0pt}
    \raggedright \scriptsize
    \textbf{Gemini-3-Pro} \newline
    $\bm{w_p}$: banana \newline
    $\bm{S_p}$: A long curved fruit which grows in clusters and has soft pulpy flesh and yellow skin when ripe. \newline
    $\bm{w_a}$: soul \newline
    $\bm{S_a}$: The spiritual or immaterial part of a human being or animal, regarded as immortal. \newline
    \textbf{Explanation}: The pun relies on the phonetic similarity between the word 'banana' and the word 'soul'. The phrase 'banana mate' sounds like 'soulmate', which is a common term for a perfect romantic partner or close friend. The image depicts two bananas hugging affectionately, reinforcing the 'mate' aspect, while the word 'banana' directly describes the visual subjects.

\end{minipage}
\\ % 行结束

% ==================== 第 3 行：Semantic Hallucination ====================
\midrule
\multicolumn{2}{l}{\color{blue} \bf Semantic Hallucination}   \\
\midrule
\bf Homophonic Pun & \bf  Random Substitution (RS)\\

% --- 左列 ---
\begin{minipage}[t]{0.28\linewidth} 
    \vspace{0pt} 
    \centering
    \includegraphics[width=\linewidth]{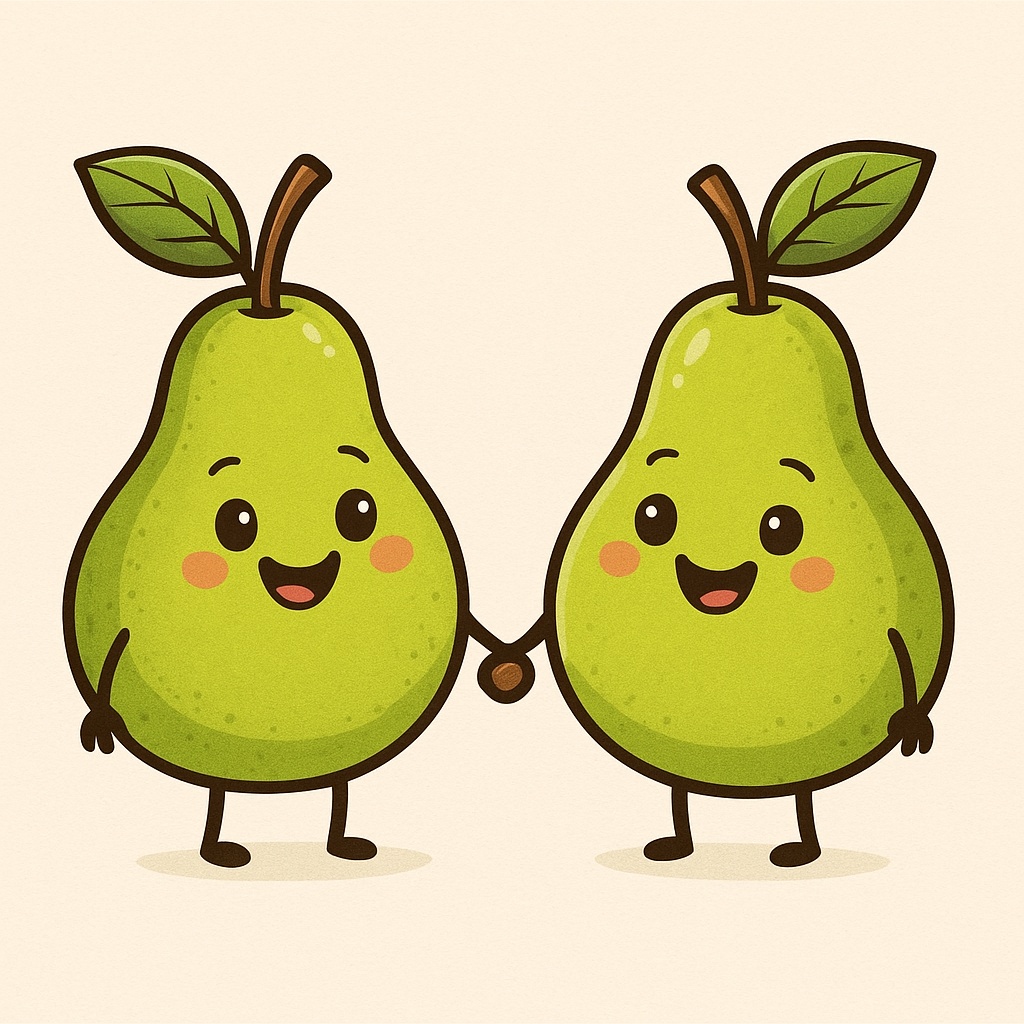} \\
    \vspace{2pt} 
    We make a great \textcolor{deepred}{\textbf{pear}}.
\end{minipage}%
\hfill 
\begin{minipage}[t]{0.68\linewidth} % 剩余空间给文字(58%)
    \vspace{0pt} % 【关键】与左侧图片顶部对齐
    \raggedright \scriptsize % 左对齐
    \textbf{GPT-4o} \newline
    $\bm{w_p}$: pear \newline
    $\bm{S_p}$: A type of fruit depicted in the image. \newline
    $\bm{w_a}$: pair \newline
    $\bm{S_a}$: Two individuals who complement each other, forming a pair. \newline
    \textbf{Explanation}: The word 'pear' in the caption sounds like 'pair'. The image depicts two pears holding hands (a visual representation of 'pear'), while the caption implies a 'pair' of individuals who complement each other.
\end{minipage}

& % --- 右列 ---
\begin{minipage}[t]{0.28\linewidth}
    \vspace{0pt}
    \centering
    \includegraphics[width=\linewidth]{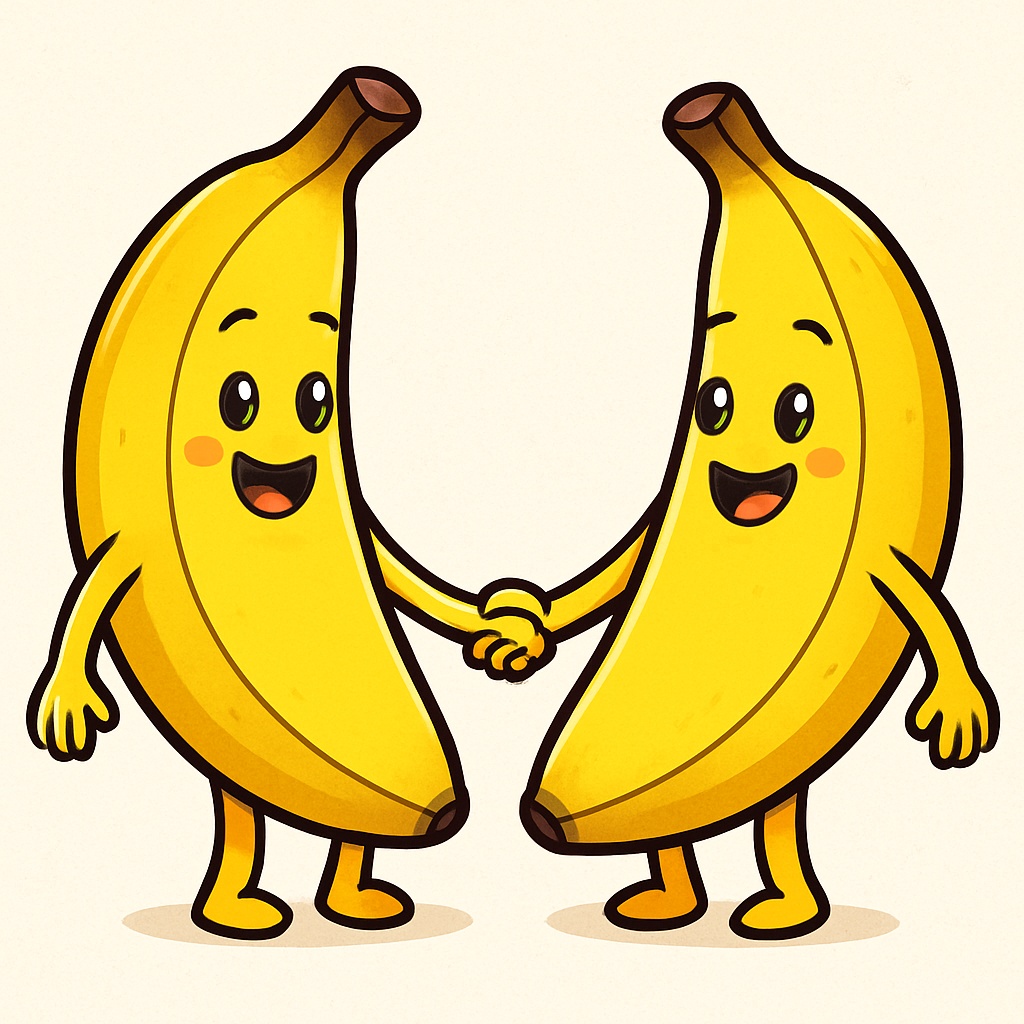} \\
    \vspace{2pt}
    We make a great \textcolor{deepred}{\textbf{banana}}. 
\end{minipage}%
\hfill
\begin{minipage}[t]{0.68\linewidth}
    \vspace{0pt}
    \raggedright \scriptsize
    \textbf{GPT-4o} \newline
    $\bm{w_p}$: banana \newline
    $\bm{S_p}$: A yellow fruit with a curved shape, as depicted in the image. \newline
    $\bm{w_a}$: banana \newline
    $\bm{S_a}$: A metaphorical reference to a pair or team that complements each other. \newline
    \textbf{Explanation}: The word 'banana' refers to the fruit depicted in the image, while in the context of the caption, it metaphorically refers to the idea of being a 'pair' or 'team' that works well together.    

\end{minipage}
\\ % 行结束

% ==================== 第 4 行：Visual Object Hallucination ====================
\midrule
\multicolumn{2}{l}{\color{blue} \bf Visual Object Hallucination}   \\
\midrule
\bf Homophonic Pun & \bf  Explicative Substitution (ES) \\

% --- 左列 ---
\begin{minipage}[t]{0.28\linewidth} 
    \vspace{0pt} 
    \centering
    \includegraphics[width=\linewidth]{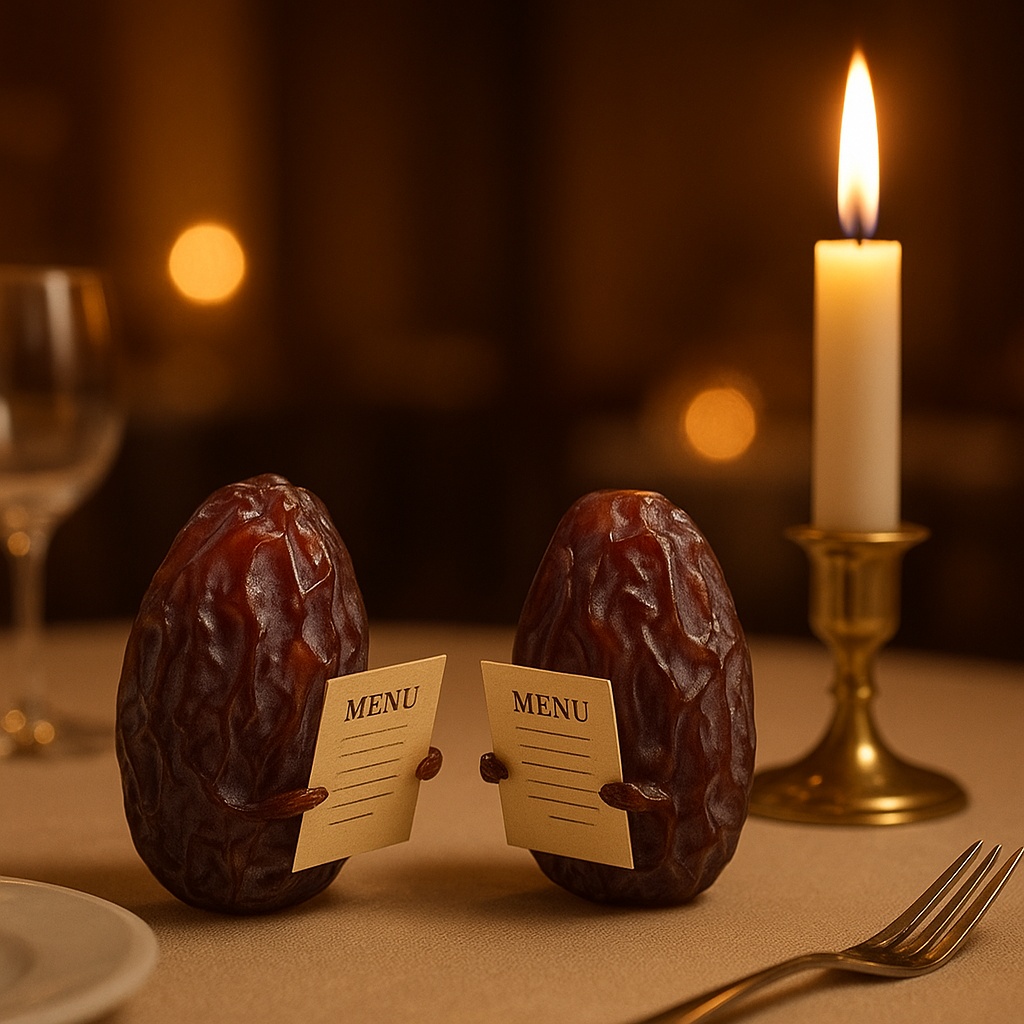} \\
    \vspace{2pt} 
    It's our first \textcolor{deepred}{\textbf{date}}.
\end{minipage}%
\hfill 
\begin{minipage}[t]{0.68\linewidth} % 剩余空间给文字(58%)
    \vspace{0pt} % 【关键】与左侧图片顶部对齐
    \raggedright \scriptsize % 左对齐
    \textbf{Qwen3-VL-30B-A3B-Instruct} \newline
    $\bm{w_p}$: date \newline
    $\bm{S_p}$: A dried fruit, specifically a date palm fruit. \newline
    $\bm{w_a}$: date \newline
    $\bm{S_a}$: A social meeting between two people, typically for romantic purposes. \newline
    \textbf{Explanation}: The word 'date' in the caption has two meanings. Visually, the image shows two dates (the fruit), which are literal 'dates'. Contextually, the phrase 'first date' refers to a romantic meeting. The humor arises from the play on the dual meaning of 'date'.

\end{minipage}

& % --- 右列 ---
\begin{minipage}[t]{0.28\linewidth}
    \vspace{0pt}
    \centering
    \includegraphics[width=\linewidth]{figures/date.jpg} \\
    \vspace{2pt} 
    It's our first \textcolor{deepred}{\textbf{romantic meeting}}. 
\end{minipage}%
\hfill
\begin{minipage}[t]{0.68\linewidth}
    \vspace{0pt}
    \raggedright \scriptsize
    \textbf{Qwen3-VL-30B-A3B-Instruct} \newline
    $\bm{w_p}$: meeting \newline
    $\bm{S_p}$: The visual representation of two dates (the fruit) as if they are on a romantic date. \newline
    $\bm{w_a}$: meeting \newline
    $\bm{S_a}$: A formal or social appointment, specifically a romantic date. \newline
    \textbf{Explanation}: The word 'meeting' in the caption has two meanings. Visually, the image shows two dates (the fruit) holding menus, which is a play on the word 'date' as in a romantic appointment. The pun lies in the word 'meeting,' which can mean a romantic date (the fruit) or a formal appointment. The humor arises from the dual meaning of 'meeting' in the context of a romantic encounter.

\end{minipage}
\\ % 行结束

\bottomrule
\end{tabular} % 结束 standard tabular
}
\caption{Qualitative examples of false positive errors on negative samples. We illustrate four distinct hallucination patterns where VLMs fail to reject non-pun distractors: (1) Pun Keyword Hallucination (Lexical Level), where models ignore visible text to restore idiomatic keywords; (2) Phonetic Hallucination (Phonological Level), where models fabricate sound similarities between distinct terms; (3) Semantic Hallucination (Semantic Level), where models force polysemous meanings onto unambiguous synonyms; and (4) Visual Object Hallucination (Visual Level), where models misidentify visual objects to align with textual cues.} 
\label{tab:appendix_negative_cases}
\end{table*}

\subsection{Error Analysis in Pun Explanation}
\label{app:appendix_positive_cases}

While VLMs can often detect puns, they struggle with providing accurate explanations. Table~\ref{tab:appendix_positive_cases} categorizes four distinct failure modes: detection failures, pun word errors, alternative word errors, and cross-modal integration errors.

\begin{table*}[h]
\centering
\footnotesize 
\definecolor{deepred}{RGB}{200, 0, 0}

\setlength{\belowrulesep}{1pt} 
\setlength{\aboverulesep}{1pt}
\resizebox{\linewidth}{!}{
\begin{tabular}{p{0.13\textwidth} p{0.48\textwidth} p{0.39\textwidth}}
\toprule

% ======================================================================
\multicolumn{3}{l}{\color{blue} \bf Detection Failure}   \\
\midrule
\begin{minipage}[t]{\linewidth}
    \vspace{0pt} 
    \centering
    \includegraphics[width=\linewidth]{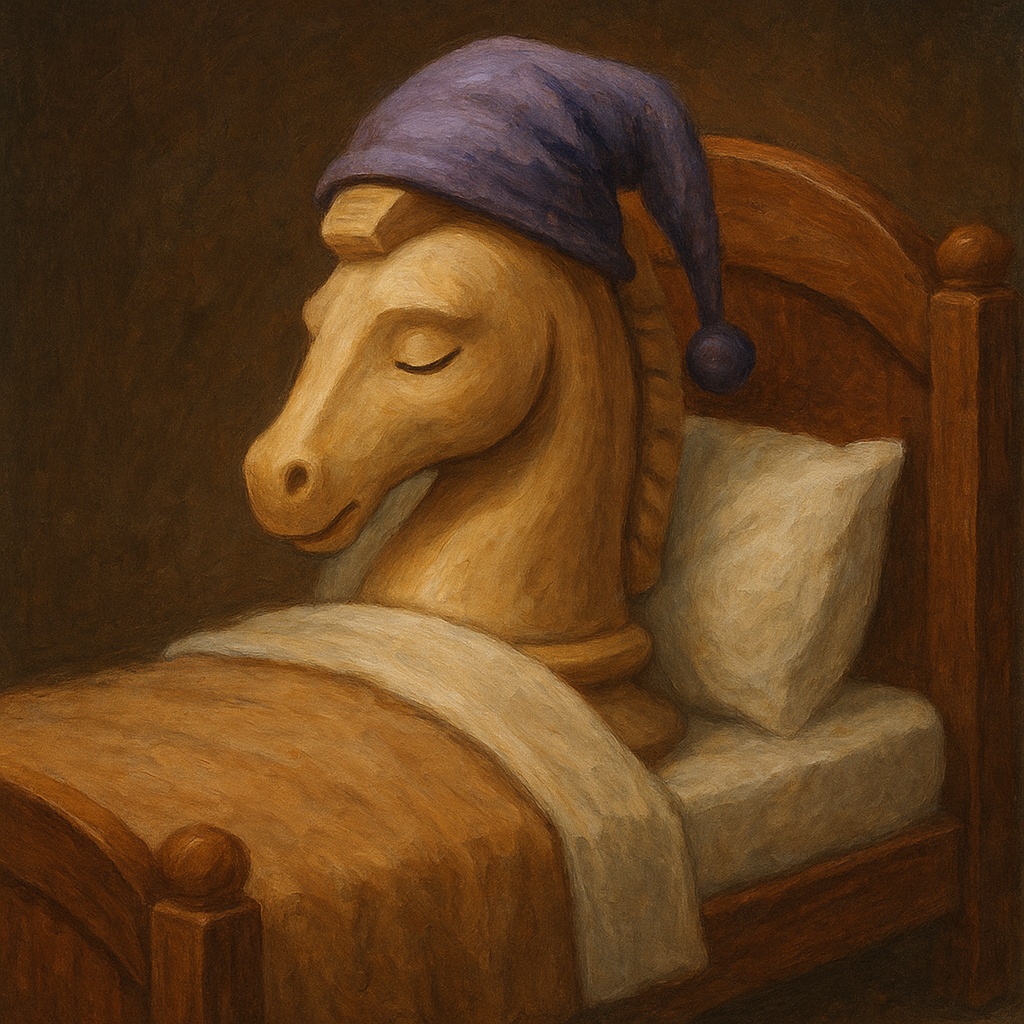} \\
    \vspace{2pt}
    Good \textcolor{deepred}{\textbf{knight}}.
\end{minipage} &
\begin{minipage}[t]{\linewidth}
    \vspace{0pt} 
    \raggedright 
    \textbf{Homophonic Pun}  \newline
    \vspace{2pt} 
    $\bm{w_p}$: knight \newline
    $\bm{S_p}$: originally a person of noble birth trained to arms and chivalry; today in Great Britain a person honored by the sovereign for personal merit \newline
    $\bm{w_a}$: night \newline
    $\bm{S_a}$: the time after sunset and before sunrise while it is dark outside \newline
    \textbf{Explanation}: The visual shows a chess knight piece wearing a nightcap and lying in bed, symbolizing bedtime. The caption, "Good knight," creates a pun by substituting "knight" (the chess piece) for "night" in the common bedtime farewell, "Good night." The humor arises from the homophonic wordplay, as "knight" and "night" are pronounced the same but have different spellings and meanings.
\end{minipage} &
\begin{minipage}[t]{\linewidth}
    \vspace{0pt} 
    \raggedright 
    \textbf{Qwen3-VL-8B-Instruct} \newline
    \vspace{2pt} 
    is pun: False \newline
\end{minipage} \\

% ======================================================================
\midrule
% ======================================================================
\multicolumn{3}{l}{\color{blue} \bf Pun Words Error}   \\
\midrule
\begin{minipage}[t]{\linewidth}
    \vspace{0pt} 
    \centering
    \includegraphics[width=\linewidth]{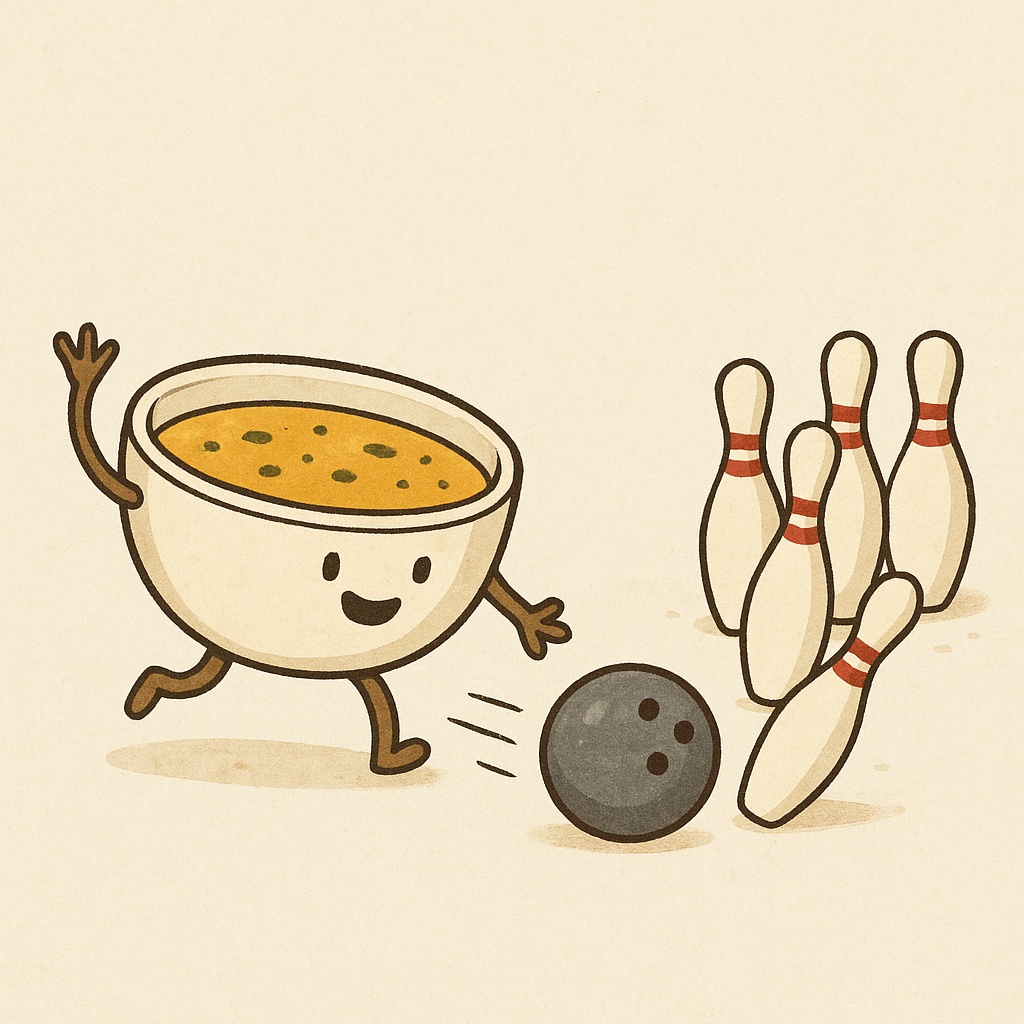} \\
    \vspace{2pt}
    Super \textcolor{deepred}{\textbf{bowl}}.
\end{minipage} &
\begin{minipage}[t]{\linewidth}
    \vspace{0pt} 
    \raggedright 
    \textbf{Homophonic Pun} \newline
    \vspace{2pt} 
    $\bm{w_p}$: bowl \newline
    $\bm{S_p}$: a round vessel that is open at the top; used chiefly for holding food or liquids \newline
    $\bm{w_a}$: bowl \newline
    $\bm{S_a}$: a wooden ball (with flattened sides so that it rolls on a curved course) used in the game of lawn bowling \newline
    \textbf{Explanation}: The visual features a soup bowl (dish) depicted as engaging in the sport of bowling, humorously anthropomorphized to throw a ball at pins. The caption, "Super bowl," creates a pun through homography, as it plays on the identical spelling of "bowl" (a dish) and "bowl" (a sporting action). The humor arises from the unexpected juxtaposition of a kitchen item with sports terminology.

\end{minipage} &
\begin{minipage}[t]{\linewidth}
    \vspace{0pt} 
    \raggedright 
    \textbf{gemini-3-pro} \newline
    \vspace{2pt} 
    $\bm{w_p}$: super \newline
    $\bm{S_p}$: An adjective describing the bowl of soup depicted in the image, specifically highlighting its identity as 'soup'. \newline
    $\bm{w_a}$: soup-er \newline
    $\bm{S_a}$: An adjective meaning excellent or of high grade; also part of the proper noun 'Super Bowl' referring to the NFL championship game. \newline
    \textbf{Explanation}: The humor relies on the phonetic similarity between the phrase 'Super bowl' (referring to the major American football championship event) and the phrase 'soup-er bowl' (describing a bowl of soup that is excellent at bowling). The image literally depicts a bowl of soup bowling, creating a visual pun on the homophone.
\end{minipage} \\

% ======================================================================
\midrule
% ======================================================================
\multicolumn{3}{l}{\color{blue} \bf Alternative Word Error}   \\
\midrule
\begin{minipage}[t]{\linewidth}
    \vspace{0pt} 
    \centering
    \includegraphics[width=\linewidth]{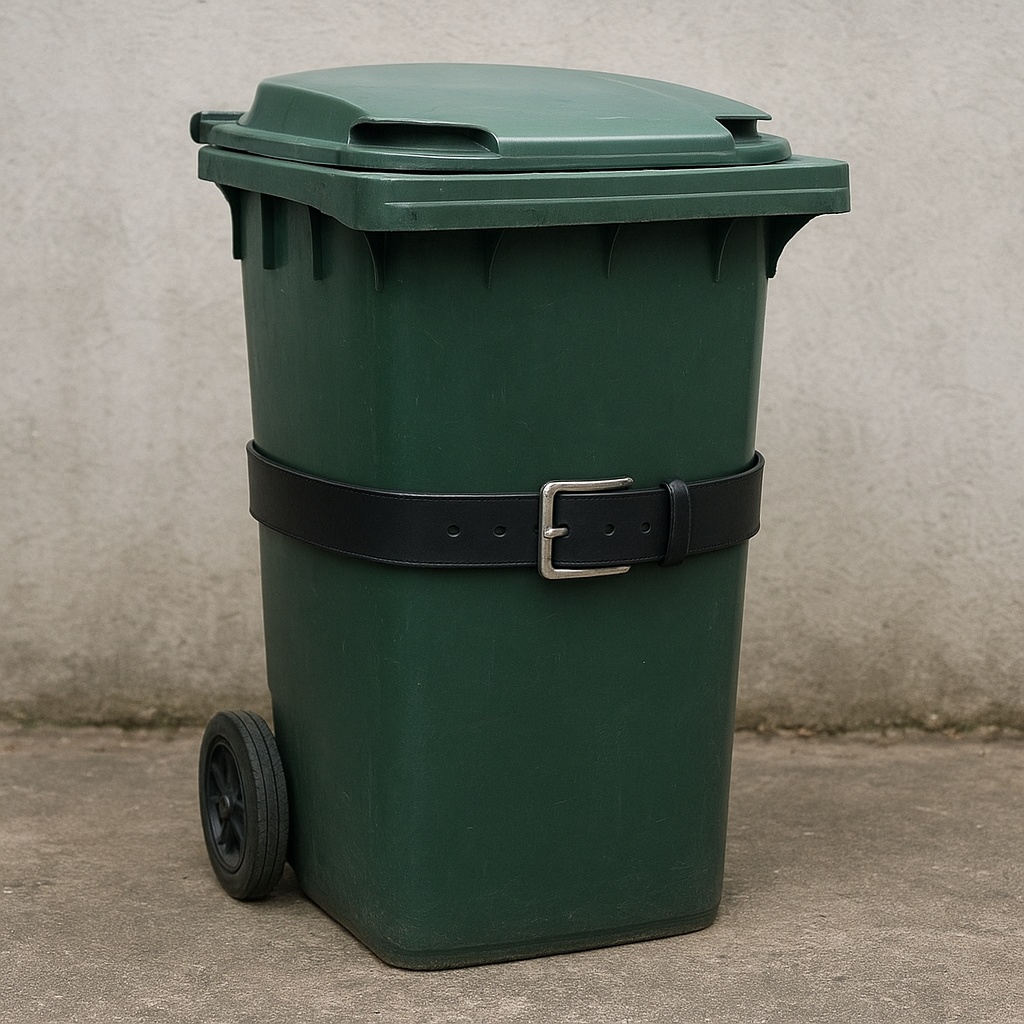} \\
    \vspace{2pt}
    What a \textcolor{deepred}{\textbf{waist}}.
\end{minipage} &
\begin{minipage}[t]{\linewidth}
    \vspace{0pt} 
    \raggedright 
    \textbf{Homophonic Pun} \newline
    \vspace{2pt} 
    $\bm{w_p}$: waist \newline
    $\bm{S_p}$: the narrowing of the body between the ribs and hips \newline
    $\bm{w_a}$: waste \newline
    $\bm{S_a}$: any materials unused and rejected as worthless or unwanted \newline
    \textbf{Explanation}: The visual shows a garbage bin (representing "waste") wearing a belt around its middle, resembling a human waist. The caption, "What a waist," creates a homophonic pun by playing on the identical pronunciation of "waist" (a body part) and "waste" (garbage). The humor arises from the interplay between the visual of the trash bin and the double meaning of the caption.
\end{minipage} &
\begin{minipage}[t]{\linewidth}
    \vspace{0pt} 
    \raggedright 
    \textbf{Qwen3-VL-30B-A3B-Thinking} \newline
    \vspace{2pt} 
    $\bm{w_p}$: waist \newline
    $\bm{S_p}$: The belt wrapped around the trash bin, mimicking a human waistline \newline
    $\bm{w_a}$: waist \newline
    $\bm{S_a}$: The narrow part of the human torso between the ribs and hips, typically used in compliments about body shape \newline
    \textbf{Explanation}: The word 'waist' in the caption refers to both the belt around the trash bin (visual sense) and the human body part (textual context), creating humor through the unexpected application of a human compliment to an inanimate object. 
\end{minipage} \\

% ======================================================================
\midrule
% ======================================================================
\multicolumn{3}{l}{\color{blue} \bf Cross-modal Integration Error}   \\
\midrule
\begin{minipage}[t]{\linewidth}
    \vspace{0pt} 
    \centering
    \includegraphics[width=\linewidth]{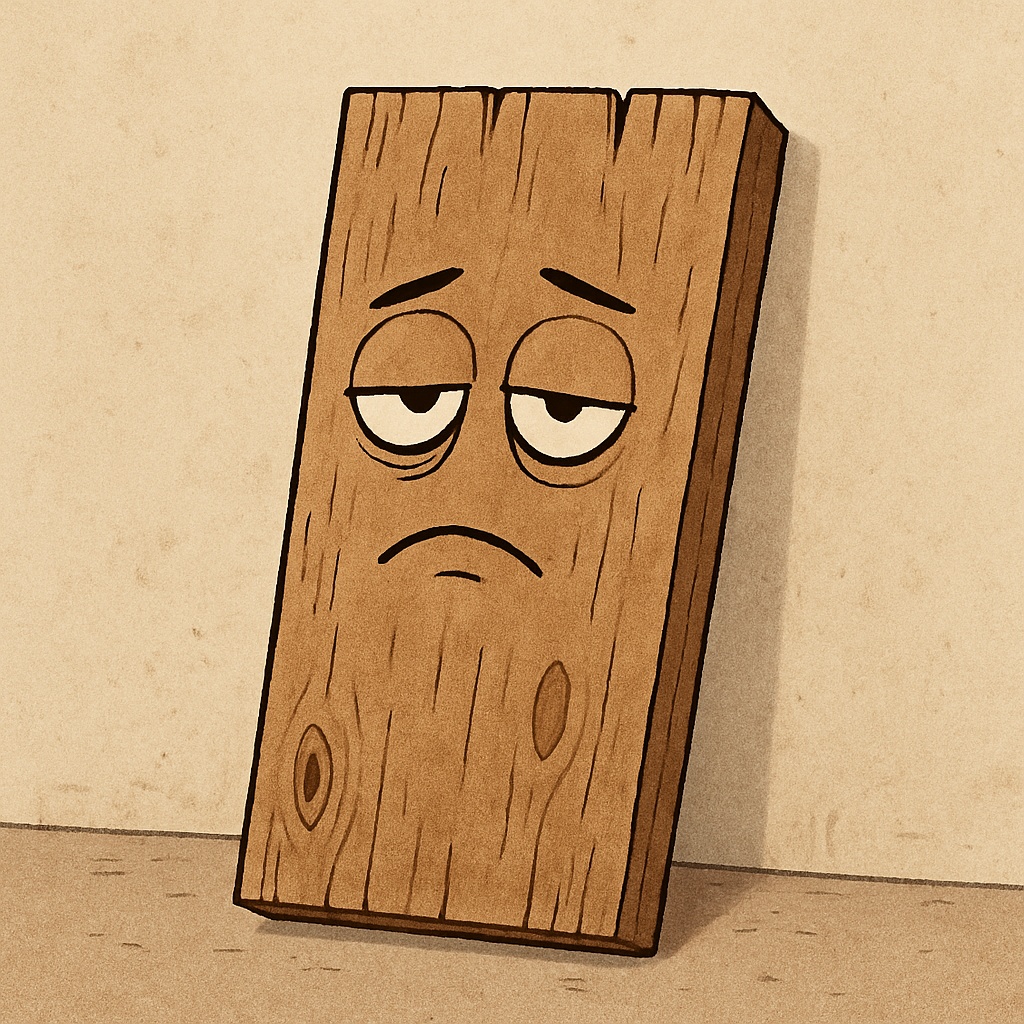} \\
    \vspace{2pt}
    I'm so \textcolor{deepred}{\textbf{board}}.
\end{minipage} &
\begin{minipage}[t]{\linewidth}
    \vspace{0pt} 
    \raggedright 
    \textbf{Homophonic Pun} \newline
    \vspace{2pt} 
    $\bm{w_p}$: board \newline
    $\bm{S_p}$: a flat piece of material designed for a special purpose \newline
    $\bm{w_a}$: bored \newline
    $\bm{S_a}$: uninterested because of frequent exposure or indulgence \newline
    \textbf{Explanation}: The visual features a wooden plank (a "board") depicted with a sleepy, disinterested face. The caption, "I'm so board," creates a homophonic pun by playing on the similar pronunciation of "board" (a piece of wood) and "bored" (a state of monotony). The humor arises from the dual interpretation of the word, linking the object to the emotion.
\end{minipage} &
\begin{minipage}[t]{\linewidth}
    \vspace{0pt} 
    \raggedright 
    \textbf{GPT-4o} \newline
    \vspace{2pt} 
    $\bm{w_p}$: board \newline
    $\bm{S_p}$: A flat, thin, rectangular piece of material, such as wood, depicted in the image. \newline
    $\bm{w_a}$: board \newline
    $\bm{S_a}$: The state of being bored or uninterested, implied by the context of the caption. \newline
    \textbf{Explanation}: The word 'board' has two meanings: one relates to the wooden plank shown in the image, and the other refers to the feeling of boredom implied in the caption. 
\end{minipage} \\
\bottomrule
\end{tabular}
}
\caption{Qualitative examples of failures on positive samples (genuine puns). We identify four failure modes: (1) Detection Failure, where the pun is missed entirely; (2) Pun Words Error, where the model focuses on the wrong lexical trigger; (3) Alternative Word Error, where the model fails to retrieve the hidden meaning ($w_a$) of the anchor word; and (4) Cross-modal Integration Error, where the model confuses the linguistic mechanism (e.g., treating homophony as polysemy).} 
\label{tab:appendix_positive_cases}
\end{table*}

\end{document}